%% file: main.tex
\documentclass{article}

\PassOptionsToPackage{numbers}{natbib}

    \usepackage[preprint]{neurips_2023}

\usepackage[utf8]{inputenc} 
\usepackage[T1]{fontenc}    
\usepackage{url}            
\usepackage{booktabs}       
\usepackage{multirow}
\usepackage{amsfonts}       
\usepackage{nicefrac}       
\usepackage{microtype}      
\usepackage{xcolor}         

\usepackage{amsmath,amssymb}
\usepackage{amsthm}
\usepackage{xcolor}
\usepackage{wrapfig}

\usepackage{newfloat}
\usepackage{listings}
\usepackage{diagbox}
\usepackage{algorithm}
\usepackage{algpseudocode}
\usepackage{amsthm}
\usepackage[english]{babel}

\usepackage{hhline}
\usepackage{subcaption}

\usepackage{enumitem}
\usepackage{graphicx}
\usepackage{hyperref}

\usepackage{tikz}
\usepackage{times}
\usepackage{pbox}
\usepackage{mathtools} 
\usepackage{enumitem}
\usepackage{amsxtra}
\newtheorem{theorem}{Theorem}
\newtheorem{lemma}{Lemma}
\newtheorem{Corollary}{Corollary}

\newcommand{\var}{{\operatorname{Var}}}
\newcommand{\OIS}{\operatorname{OIS}}

\title{Statistically Efficient Variance Reduction with Double Policy Estimation for Off-Policy Evaluation in Sequence-Modeled Reinforcement Learning} 

\author{%
  Hanhan Zhou \\
    The George Washington University\\
  \texttt{hanhan@gwu.edu} \\
  \And
  Tian Lan \\
  The George Washington University\\
  \texttt{tlan@gwu.edu} \\
  \AND
  Vaneet Aggarwal \\
  Purdue University \\
  \texttt{vaneet@purdue.edu} \\
}

\begin{document}

\maketitle

\begin{abstract}
Offline reinforcement learning aims to utilize datasets of previously gathered environment-action interaction records to learn a policy without access to the real environment. Recent work has shown that offline reinforcement learning can be formulated as a sequence modeling problem and solved via supervised learning with approaches such as decision transformer. While these sequence-based methods achieve competitive results over return-to-go methods, especially on tasks that require longer episodes or with scarce rewards, importance sampling is not considered to correct the policy bias when dealing with off-policy data, mainly due to the absence of behavior policy and the use of deterministic evaluation policies. To this end, we propose DPE: an RL algorithm that blends offline sequence modeling and offline reinforcement learning with Double Policy Estimation (DPE) in a unified framework with statistically proven properties on variance reduction. We validate our method in multiple tasks of OpenAI Gym with D4RL benchmarks. Our method brings a performance improvements on selected methods which outperforms SOTA baselines in several tasks, demonstrating the advantages of enabling double policy estimation for sequence-modeled reinforcement learning.
\end{abstract}

\input{1Intro}

\input{2Background}

\input{4Method}

\input{5Experiments}

\input{6Conclusion}

{
\small
\bibliography{neuripsbib}
\bibliographystyle{unsrtnat} 
}
\newpage

\input{7appendix}







\end{document}

%% file: 1Intro.tex
\section{Introduction}
Many real-world reinforcement learning (RL) problems, such as autonomous vehicle coordination and data-driven problems\cite{luo2022multisource,He2022Robust,HUA2023121526} are widely used to solve sequential decision-making problems naturally in a way as the agent takes an action based on its observation, receives a reward from the environment, and then observes the next action\cite{ma2020statistical,chen2023minimizing,gogineni2023accmer}, and so on \cite{sutton2018reinforcement,chen2021deepfreight,zhao2022implicit}. Usually modeled as a Markov Decision Process (MDP), the agent would take an action solely based on its current state information (which represents the whole trajectory history), so a scheme where sequences are divided into each step and later solved with algorithms like Temporal Difference learning (TD-learning) \cite{sutton1988learning} is proposed. This could be derived via Bellman Equations to solve RL problems mathematically.
Recent advances in offline reinforcement learning (RL) algorithms provide a promising approach for sequential decision-making tasks without the need for online interactions with an environment \cite{janner2021offline,zhang2023self}. This approach is particularly appealing when online interactions are costly or when there is an abundance of offline experiences available. Recent works have demonstrated that generative models \cite{chen2020generative,brown2020language,radford2018improving} that are widely used in language and vision tasks can be applied to maximize the likelihood of trajectories in an offline dataset without temporal difference learning \cite{Janner2021ReinforcementLA}, notably, Decision Transformer (DT) \cite{chen2021decision}, which uses the transformer architecture \cite{vaswani2017attention} for decision-making. Such a pertaining paradigm in a supervised learning manner for RL can be considered known as Reinforcement learning via Supervised Learning (RvS) \cite{rvs, Schmidhuber2019ReinforcementLU, Srivastava2019TrainingAU}. Instead of learning a value-based algorithm for decision-making, RvS-based methods often consider the learning task as a prediction problem: to predict an action that will lead to a certain outcome or reward when given a sequence of past states and actions (e.g., using causal transformer architectures). These methods have gained significant attention due to their algorithmic and implementation simplicity while bringing a robust performance on several offline-RL benchmarks. 

Learning an RvS policy $\pi_e$ requires off-policy learning since we need to estimate the expected return of the learned policy $\pi_e$ during training, from offline experiences/trajectories that are generated using a different behavior policy $\pi_b$. 
%
%
We note that online policy evaluation is usually expensive, risky, or even unethical for many real-world problems \cite{Jiang2015DoublyRO}. When the actual environment is not accessible, these trajectories sampled by $\pi_b$ can be used to evaluate $\pi_t$, also known as off-policy evaluation (OPE) \cite{sutton2018reinforcement}. An accurate OPE is crucial to evaluate and optimize a policy during training from offline datasets, the concept of importance sampling (IS) rectifies the discrepancy between the distributions of the behavior policy $\pi_b$ and the evaluation policy $\pi_e$~\cite{Precup2000EligibilityTF}. IS-based off-policy evaluation methods have also seen lots of interest recently, especially for short-horizon problems \cite{hirano2003efficient,murphy2001marginal}, including contextual bandits \cite{wang2017optimal}. However, the application of IS to sequence modeling-based RvS methods is difficult due to a number of challenges. The behavior policies for collecting experience/trajectory data are often not available, while the evaluation policies in RvS methods are typically deterministic, making reweighting different experiences/trajectories inaccessible. Further, the variance of IS-based approaches tends to be too high to provide informative results, for long-horizon problems, since the variance of the product of importance weights may grow exponentially as the horizon goes long\cite{gottesman2019combining, hanna2019importance}. 


Although it is intuitively to assume that replacing the behavior policy with its empirical estimation can harm the performance and increase the variance of a policy, recent works in several domains including multi-armed bandits\cite{hanna2018towards} and off-policy evaluation \cite{gogineni2023accmer,chen2023explainable,mei2023remix} have shown that by applying an estimation of the behavior policy could improve the mean squared error of importance sampling policy evaluation \cite{li2015toward}.

In this paper, we study a problem that when given a dataset of trajectories sampled by a behavior policy and trajectories generated with sequence-modeling-based evaluation policy (in this paper we select Decision Transformer to demonstrate our approach), to estimate both behavior policy and target policy and then compute the importance sampling estimate which we call double policy estimation importance sampling. We further provide a theoretical analysis on the properties of such estimators and show that this double policy estimation will reduce the variance of the target policy learned.

Specifically, we propose to introduce an asymptotic estimation for both behavior policy $\pi_b$, which is used to sample and generate the dataset, and target evaluation policy, $\pi_t$, which is the policy we are in an attempt to learn and correct, as double policy estimation, to calculate the likelihood ratio for all state-action pairs in the off-policy data. Although it may seem that such an estimation would bring even worse performance as it introduces more uncertainties\cite{he2023robust,he2020data}, recent research in several domains including multi-armed bandits \cite{li2015toward,narita2019efficient}, Monte Carlo integration \cite{delyon2016integral}, and causal inference \cite{hirano2003efficient} has shown this estimating behavior could potentially improve the mean squared error of importance sampling policy evaluation which partially motivates this design. Another direct motivation is that specifically for many generation models based RvS methods like decision transformer in an offline reinforcement a common scenario is that both $\pi_b$ and $\pi_t$ are both inaccessible, which promotes a design for double policy estimation. 

We prove that DPE can statistically lower the mean squared error of importance sampling OPE with lower variance.  We implement the proposed DPE on D4RL environments and compare DPE with SOTA baselines including DT~\cite{chen2021decision}, RvS~\cite{rvs}, CQL~\cite{cql}, BEAR~\cite{bear}, UWAC~\cite{uwac}, BC~\cite{brac}, and IQL~\cite{iql}. We empirically found double policy estimation based on importance sampling also brings an improvement to the off-policy evaluation of the D4RL environment, where DPE achieves better performance than the original decision transformer on almost all datasets and outperforms the state-of-the-art baselines over several datasets
with further analysis discussing the effects and properties of the proposed double policy estimator. 


%% file: 2Background.tex
\section{Background}

\subsection{Markov Decision Process and Sequence-Based Method in Reinforcement Learning}

We assume that the environment is a Markov decision process with a finite horizon and episodic nature, where the state space is denoted as $\mathcal{S}$, the action space as $\mathcal{A}$, and the environment possesses transition probabilities represented by $P$, a reward function denoted as $R$, a horizon length of $H$, a discount factor of $\gamma$, and initial state distribution of $d_0$ \cite{puterman2014markov,li2015toward}. 
A policy, denoted as $\pi$, is considered Markovian if it maps the current state to a probability distribution over actions. In contrast, a policy is classified as non-Markovian if its action distribution is dependent on past actions or states \cite{mei2022bayesian,agarwal2022multi}.
We assume  $\mathcal{S}$ and $\mathcal{A}$ are finite for simplicity and  probability distributions are probability mass functions.

In off-policy policy evaluation, we are given a fixed {evaluation policy}, $\pi_e$, and a data set of $m$ trajectories and the policies that generated them: $\mathcal{D}  \{\omega^i, \pi_b^{(i)}\}_{i=1}^{m}$ 
where $\omega^i \sim \pi_b^{(i)}$.
We assume that $\forall \{\omega^i, \pi_b^{(i)}\} \in \mathcal{D}$, $\pi_b^{(i)}$ is Markovian,  i.e., actions in $\mathcal{D}$ are independent of past states and actions gave the immediately preceding state \cite{ma2022traffic}. Sequence-based methods in reinforcement learning, which is trained in reinforcement learning via supervised learning (RvS) manner such as Decision Transformer, train a model using supervised learning on a dataset with respect to trajectories to predict $p_{\mathcal{D}(a|s, R)}$, i.e., given a cumulative reward $R = \sum_{t} \gamma^t r_t$ to predict the probability of next action conditioning the current state. Then at the deployment stage, the model takes actions conditioned on a desired target return value. 
Our goal is to design an off-policy estimator that takes $\mathcal{D}$ as input and estimates both behavior policy $\pi_b$ and evaluation policy $\pi_e$ for enabling importance sampling in sequence modeling methods.

Decision Transformer processes a trajectory $\boldsymbol{\omega}$ as a sequence consisting of 3 types of input to be tokenized: the states, actions selected, and the return-to-go \cite{vaswani2017attention,chen2021decision}. Specifically, it learns a deterministic model $\pi_{\text{DT}(a_t|  \textbf{a}_{-K,t}, \textbf{s}_{-K,t}, \textbf{r}_{-K,t})}$ where $-K$ denotes the past K sequences and is trained to predict the action token at timestamp $t$. During the evaluation, DT is given a desired reward $g_0$ and the initial stage $s_0$ at the beginning and executes the action it generates. Once an action $a_t$ is generated and then executed, the next state $s_{t+1} \sim P(\cdot|s_t,a_t)$ and reward ${r_t = R(s_t,a_t)}$ are observed, together with the return-to-go $g_{t+1} = r_t - g_t$: this new sequence will be appended to the previous input. The process is repeated until the terminal state. DT is then trained under standard $l_2$ loss as $\nabla_{\theta} J(\pi_{DT}) = \frac{1}{K}\sum_{k} \nabla_{\theta_{DT}} (a_k - \hat{a})^2$ in a supervised learning way.

\subsection{Importance Sampling in Reinforcement Learning}
Importance Sampling (IS) is a method for reweighting returns generated by a behavior policy 
$\pi_b$ to produce an unbiased estimate of the returns for the evaluation policy.
To obtain a reliable numerical integration of $f$ as $\theta =  \int f(x)dx$ , assuming there is a family of sampling distributions, $p(x;\eta)$, with parameter $\eta$, that generates a random trajectory $\omega := (s_0, a_0, r_0, \cdots, s_{L-1}, a_{L-1}, r_{L-1})$ from $p(x;\eta_0)$, where $g(\omega) := \sum_{t=0}^{L-1}\gamma^t r_t$ be the discounted return with preliminary fixed $\eta_0$: an ordinary importance sampling (OIS) method provides an estimator of $\theta$ in the form of $\Tilde{\theta} = \frac{1}{n} \sum_{i= 1}^{n} \frac{f(x_i)}{p(x;\eta_0)}$. Then $\Tilde{\theta}$ is an unbiased estimator of $\theta$ and $\Tilde{\theta}$ is guaranteed to converge to $\theta$ as $n$ goes to infinity according to the strong law of large numbers \cite{henmi2007importance}.

In Monte Carlo problems with high-dimensional $x$, the target density $p(x)$ can be writing in a chain-like decomposition as $p(x) = p(x_1)\prod_{t=2}^{d} p(x_1 | x_{1:t-1})$, where $x_{[1:t]} = (x_1, \cdots, x_t)$.
With a set of $m$ trajectories and the policy that generated each trajectory, the IS off-policy estimate of $v(\pi_e)$ is:
$
\operatorname{IS}(\pi_e, \mathcal{D}) \coloneqq \frac{1}{m}\sum_{i=1}^{m} g(\omega^{(i)}) \prod_{t=0}^{L-1} \frac{\pi_e(a_t^{(i)} | s_t^{(i)})}{\pi_b(a_t^{(i)} | s_t^{(i)})}.$
We refer to this as the ordinary importance sampling ($\OIS$) estimator which uses the true behavior policy and refer to $\frac{\pi_e(a| s)}{\pi_b(a | s)}$ as the OIS weight for action $a$ in state $s$. A standard approach to dealing with off-policy data is to correct the policy using importance sampling (IS) by applying cumulative density ratios $\nu_{0:t}$ \cite{kallus2020statistically, hanna2018towards}. Then the policy gradient $Z(\theta)$ can be rewritten as an expectation over $p_{\pi_b}$ and further estimated using an equivalent empirical expectation. The off-policy version of the classic REINFORCE algorithm \citep{williams1992simple} recognizes
$
   Z(\theta)=\mathbb{E}[{\nu_{0:H}{\sum_{t=0}^Hr_t}{\sum_{t=0}^Hg_t}} ]
$
(recall that $E$ is understood as $\mathbb{E}_{p_{\pi^b}}$)
and uses the estimated policy gradient given by replacing $\mathbb{E}$ with $\mathbb{E}_n$. Later works obtained a policy gradient in terms of Q-function as $      Z(\theta)=\mathbb{E}[\sum_{t=0}^{H}\nu_{0:t}g_t q_t]$ \cite{chen2019information}.


\section{Related Work}

\subsection{Sequence-Based method in Reinforcement Learning}

Much recent progress has been on formulating the offline decision-making procedure in offline reinforcement learning as a context-conditioned sequence modeling problem \cite{Janner2021ReinforcementLA,chen2021decision}. Compared to the temporal difference methods, these works consider a paradigm that utilizes predictive models to generate desired actions from the observation sequence and the task specification like a supervised learning problem \cite{Schmidhuber2019ReinforcementLU, Srivastava2019TrainingAU, rvs} rather than learning a Q-function or policy gradients. Specifically, the Decision Transformer model \cite{chen2021decision} trains the transformer architecture \cite{vaswani2017attention} as a model-free context-conditioned policy that takes the encoded reward-to-go, state, and action sequence as input to predict the action for the next step, and the Trajectory Transformer \cite{Janner2021ReinforcementLA} trains transformer that first discretizes each dimension of the input sequence and shows that beam search can be used to improve upon the model-free performance. Various attempts have also been made to improve transformers in multi-agent RL \cite{zhou2022pac,chen2022scalable, mei2023mac, zhou2022value} and other areas including meta RL \cite{chen2021bringing,chen2023minimizing}, and multi-task RL\cite{chen2023option}.
However, these works do not consider the importance of sampling for offline reinforcement learning. Our work extended this area with the proposed double policy estimation and further improved the asymptotic variance of the ordinary method using the true sampling distribution.

\subsection{Importance Sampling in Reinforcement Learning}
The use of off-policy samples within reinforcement learning is a popular research area \cite{silver2014deterministic,levine2013guided,elmachtoub2023balanced}. Many of them rely on OIS or variants of OIS to correct for bias. The use of importance sampling ensures unbiased estimates, but at the cost of considerable variance, as quantified by the ESS measure \cite{jie2010connection}. The problem of sampling error applies to any variant of importance sampling using OIS weights, e.g., weighted importance sampling and per-decision importance sampling \cite{Precup2000EligibilityTF}, the doubly robust estimator \cite{Jiang2015DoublyRO}, and the MAGIC estimator \cite{Thomas2016DataEfficientOP}. On-policy Monte Carlo policy evaluation is also subject to sampling error, as it is a specific case of ordinary importance sampling where the behavior policy and the evaluation policy are identical. Among these important sampling methods, \cite{hanna2019importance} is the closest work but considers estimated behavior policy where their behavior policy estimate comes from the same set of data used to compute the importance sampling estimate; while we estimate the behavior policy to the training phase from the dataset and estimate the target policy from data generated from the target policy.


 

%% file: 4Method.tex
\section{Methodology}
In this section, we present the primary focus of our work: double policy estimation (DPE) importance sampling  that corrects for sampling error in sequence modeling-based reinforcement learning. The key idea is to obtain the maximum likelihood estimate of both behavior and evaluation policies $\hat{\pi}_b^\eta$ and $\hat{\pi}_t^\psi$ and use them for computing the DPE cumulative density ratio. We further analyze the theoretical properties of DPE and prove that it is guaranteed to reduce the asymptotic variance of policy parameters. A table of key notations with explanations is summarized in the Appendix.


\subsection{DPE for sequence modeling-based reinforcement learning}

Let $\mathcal{D}$ be a set of off-policy trajectories of length $H+1$ collected by a behavior policy $\pi_b$, denoted by $\mathcal{D}=\{\omega_i, \ \forall i \}$ with each trajectory ${\omega}_i= \{({s_0}^{(i)}, {a_0}^{(i)}, {r_0}^{(i)}, \cdots, {s_H}^{(i)}, {a_H}^{(i)}, {r_H}^{(i)})$). 
For known behavior policy $\pi_b$ and evaluation policy $\pi_e^{\theta}$, OIS leverages the cumulative density ratio $\nu_{0:t}=\prod_{k=0}^t v_k$ (with density ratio $v_k={\pi_e^\theta(a_k|s_k)} / {\pi_b(a_k|s_k)}$) to reweight the policy scores $g_t=\nabla_\theta \log \pi_e^\theta(a_t|s_t)$, such that they are unbiased estimates of the evaluation policy $\pi_e^\theta$. In the off-policy version of the classic REINFORCE algorithm~\cite{williams1992simple}, the policy gradient under OIS is recognized as 
$ Z(\theta)=\mathbb{E}[{\nu_{0:H} q_{0:H} {\sum_{t=0}^Hg_t}} ] $, where $q_{t:H}=\sum_{s=t}^H r_s$ is the return-to-go from step $t$ to step $H$ in trajectory $\omega$ (generated from  behavior policy $\pi_b$). OIS can be easily extended to its step-wise form
~\cite{deisenroth2013survey,chen2019information} with $ Z(\theta)=\mathbb{E}[\sum_{t=0}^H {\nu_{0:t} q_{t:H} g_t} ]$. OIS has been commonly used in off-policy reinforcement learning.




We note that when RL is recast as an offline sequence modeling problem (such as Decision Transformer \cite{chen2021decision} and RvS \cite{rvs}), it also relies on off-policy learning. However, there are three challenges preventing OIS from being directly applied to sequence modeling-based RL. First, offline RL datasets often do not provide the actual behavior policy for collecting trajectories, making it impossible to access $\pi_b$ in importance sampling. Second, sequence modeling-based RL usually are trained using a transformer structure to represent evaluation policy and to generate deterministic action outputs \cite{chen2021decision}. We need to extend them to stochastic policies to obtain $\pi_e$ in importance sampling.
Finally, OIS is known to have a high variance \cite{rasmussen2003bayesian}, also known as high sampling error in importance sampling\cite{hanna2019importance}. Methods to reduce importance-sampling variance are needed for sequence modeling-based RL.


To this end, we propose two maximum likelihood estimators of (stochastic) behavior and evaluation policies in sequence modeling-based RL, denoted by $\hat{\pi}_b^\eta$ and $\hat{\pi}_e^\psi$. A baseline return $b_t^{\xi}$ is further estimated (using a mean-square error loss) in sequence modeling-based RL and is leveraged to mitigate the variance in policy learning. Given a set $\mathcal{D}$ of $m$ trajectories, the proposed DPE with respect to the off-policy version of classic REINFORCE algorithm~\cite{williams1992simple} is defined as:
\begin{eqnarray} \label{dpe_e}
Z_{\text{DPE}}(\theta|\eta,\psi,\xi, \mathcal{D}) 
= \mathbb{E} \left[
\left( q_{0:H}-b_0^{\xi} \right) \prod_{t=0}^{H}  \frac{\pi_e^\psi(a_t|s_t)}{\pi_b^\eta(a_t|s_t)} \left( \sum_{t=0}^H g_t \right) \right].
\end{eqnarray}
DPE can also be applied to the step-wise  
form~\cite{deisenroth2013survey, chen2019information}, by replacing the density ratio ${v}_k$ with its estimator $\hat{v}_k={\pi_e^\theta(a_k|s_k)} / {\pi_b(a_k|s_k)}$ and by subtracting the return baseline $b_t^\xi$, i.e.,
\begin{eqnarray}
\label{dpe_s}
Z_{\text{DPE}}(\theta|\eta,\psi,\xi, \mathcal{D}) 
= \mathbb{E} \left[
\sum_{t=0}^H (q_{t:H}-b_t^\xi) \hat{v}_{0:t}  g_t \right].
\end{eqnarray}

The key idea of our DPE estimator for importance sampling is to leverage the maximum likelihood estimate of behavior and evaluation policies, denoted by $\hat{\pi}_b^\eta$ and $\hat{\pi}_t^\psi$ respectively. We introduce the proposed maximum likelihood estimators for $\hat{\pi}_b^\eta$ and $\hat{\pi}_e^\psi$ and minimum-mean-square estimator for $b^{\xi}$ as following:

\paragraph{Maximum likelihood estimator for behavior policy $\hat{\pi}_b^\eta$.} We consider estimating the $\hat{\pi}_b$ , with maximum likelihood as 
$
\hat{\pi}_b^\eta := \text{argmax}_{\pi_b} \sum_{\omega \in \mathcal{D}} \sum_{t} \text{log} \pi_b(a| \omega_{t-n:t})
$, so that it could provide a behavior policy action probability estimation while the training of DT. Specifically, in this work, for policy network estimator we consider learning $\pi_b$ from $\mathcal{D}$ as a Gaussian distribution over actions with mean and standard deviation estimated from a neural network.

\paragraph{Maximum likelihood estimator for target policy $\hat{\pi}_t^\psi$.} One key insight in this paper is that when assuming a Gaussian policy for target policy estimation, the estimator would be minimizing the mean-square error of action predictions, thus it is identical to sequence modeling-based RL like DT with MSE loss where its variance is this MSE specifically to each timestep while training. When obtaining the target policy estimator, although for decision transformer $\pi_b$ is often not directly available and $\pi_b(a|s, R)$ cannot be used as this estimator, also estimating an ongoing learning method might be unstable and inefficient, we point out that this weight at specific timestep $t$ can be considered as a Gaussian distribution with a mean of $\hat{a}_t$ and variance of the corresponding MSE. We explain why this can serve as target policy estimation later in the main theorem in detail.

\paragraph{Minimum-mean-square estimator for baseline $b^{\xi}$.} Since $b^{\xi}$ is trained to predict return-to-go by minimizing loss $\sum_{i=1}^m  \left[q_{t:H} - b_t^\xi \right]^2$. This can be easily incorporated into sequence modeling-based Reinforcement Learning like Decision Transformer.

\paragraph{Training sequence modeling based RL using DPE.} We summarize the general architecture of the learning pipeline on Algorithm 1 of applying DPE to the sequence-modeling-based target policy (Decision Transformer). We first obtain an empirical estimator of the behavior policy $\pi_b$ prior to the training of the Decision Transformer in a warm-up phase. Then during the training phase, we acquire the target policy estimator as a Gaussian distribution $\hat{a^\eta_t} \sim \mathcal{N}(\hat{a_t}, \sigma^2)$ where $\hat{a_t}$ is the mean generated from the decision transformer, $\hat{\sigma^2}$ is the MSE that serves as variance from the loss calculated at a specific timestamp. We present a pseudocode of the DPE training procedure in the appendix.
\subsection{Problem formulation and DPE Objective}
In offline sequence modeling-based reinforcement learning, we are given a data set of $m$ offline trajectories $\omega= \{(s_0, a_0, r_0 ...)\}$, and the behavior policy $\pi$ that is collected them. We denote the trajectories that are generated by the decision transformer as $\hat{\omega}= \{(\hat{s_0}, \hat{a_0}, \hat{r_0} ...)\}$


We consider the following two joint objectives: 
\begin{eqnarray}
\left\{\begin{matrix}
 H(\pi_t) = - \mathbb{E}\left\|    \sum_{t} \text{log}(2\pi e \sigma_t^2)\right\|,  
\\

 \\
L = - E_{\pi_t} \text{log}  q(a_t)
\\
\end{matrix}\right.
\end{eqnarray}
where minimizing $L-\beta H$ for $\pi_t$, $min (H(\pi_t))$ is to approximate the target policy decision transformer and $L$ is to maximize the likelihood of $a_t$. We then choose $\pi_b (\eta )$ to maximize the likelihood and $b(\xi)$ to minimize the squared error $\sum_{i=1}^n w^2 \cdot (G_i - b(\xi))^2 $ .




Note that DPE objective can also be written as :

\begin{eqnarray}
\text{DPE} := \frac{1}{m} \sum_{i=1}^{n} q(h_t) \prod_{t=0}^{L-1} \frac{\hat{\pi}_t^{(i)}(a_t^{(i)} | s_t^{(i)})}{\hat{\pi}_b^{(i)}(a_t^{(i)} | s_t^{(i)})} = \frac{1}{m} \sum_{i=1}^{n} \frac{\hat{w}_{\pi_t}(h_t)}{\hat{w}_{\pi_b}(h_t)} q(h_t) 
\end{eqnarray}

The variance of $\Tilde{\theta}$ is given by $\delta^2(f)/n$, where $\delta^2 = \delta^2(f) = \int \{ \frac{f(x)}{p(x; \eta_0) - \theta} \}^2p(x;\eta_0)dx$, thus the distribution of $\sqrt{n}(\Tilde{\theta}-\theta)$ converges to Normal distribution $\mathcal{N}(0, \delta^2)$ as $n$ increases to infinity according to central limit theorem. 



\subsection{Theoretical Properties of DPE}

We analyze the asymptotic properties of the maximum likelihood estimator of behavior policy $\pi_b^{\hat{\eta}}$ (with optimal parameters $\hat{\eta}$), the maximum likelihood estimator of evaluation policy $\pi_e^{\hat{\psi}}$ (with optimal parameters $\hat{\psi}$), and the minimum mean-square error estimators of baseline $b_t^{\xi}$ (with optimal parameters $\hat{\xi}$). We show that
these estimators are able to reduce the variance of policy gradient estimates $Z_{\rm DPE}$. More precisely, for a given set of $m$ off-policy trajectories $\mathcal{D}=\{\omega_i, \ \forall i \}$, we consider the gradient estimate $Z_{\rm DPE}$ with DPE (in both per-episode form as~Eq. (\ref{dpe_e}) and per-step form as~Eq. (\ref{dpe_s})), i.e.,
\begin{eqnarray}
Z_{\rm DPE} = \frac{1}{m} \sum_{i=1}^m (q_{0:H}^{(i)}-b_0^{\hat{\xi}}) \hat{v}_{0:H}^{(i)} \left( \sum_{t=0}^H g_t^{(i)} \right) \ {\rm and} \ Z_{\rm DPE} = \frac{1}{m} \sum_{i=1}^m \sum_{t=0}^H (q_{t:H}^{(i)}-b_t^{\hat{\xi}}) \hat{v}_{0:t}^{(i)} g_t^{(i)}.
\end{eqnarray}
We show that the variance $\var(Z_{\rm DPE})$ using optimal estimators $\hat{\psi}$, $\hat{\eta}$ and $\hat{\xi}$ is lower than the variance $\var(Z_{\rm OIS})$ using some ground truth ${\psi}_0$, ${\eta}_0$ and ${\xi}_0$. 

We begin with recognizing that both per-episode and per-step DPE can be consolidated using a general form:
\begin{eqnarray}
Z_\text{DPE} = \frac{1}{n} \sum_{i=1}^{n} \frac{f(\omega_i; \hat{\psi})[ G(\omega_i)- b(\hat{\xi})]}{P(\omega_{i};\hat{\eta})}
\end{eqnarray}
Next, we show a few lemmas demonstrating some properties of the estimators $\hat{\psi}$, $\hat{\eta}$, and $\hat{\xi}$ and then prove the variance reduction lemma.


\begin{lemma} Let $F_\eta = -\frac{1}{m}\sum_{i=1}^m \partial^2_\eta \log P(\omega_{i};\hat{\eta_0}) $ be the Fisher Information Matrix. We have
\begin{eqnarray}
\sqrt{m} (\hat{\eta} - \eta_0) = \frac{1}{\sqrt{m}} F^{-1}_{\eta} \cdot \sum_{i=1}^{m} \partial_\eta \log P(\omega_{i};\eta_0)  + O(1) 
\end{eqnarray}
\end{lemma}
\textbf{Proof Sketch.} Since $\hat{\eta}$ is the maximum likelihood estimator that optimizes $P(\omega_{i};{\eta})$, we have $\partial_\eta \sum_{i=1}^m \log P(\omega_{i};{\eta}) =0 $ at $\eta = \hat{\eta}$. Expanding the left-hand side from $\eta = {\eta}_0$ toward $\eta = \hat{\eta}$, we have $0=\sum_{i=1}^m \partial_\eta \log P(\omega_{i};{\eta}_0) + \sum_{i=1}^m \partial^2_\eta \log P(\omega_{i};\hat{\eta_0}) \cdot(\hat{\eta} -\eta_0) + o(||\hat{\eta} -\eta_0 ||_2)$, which yields the desired result by rearranging the terms and leveraging Fisher Information Matrix $F_\eta$.

\begin{lemma} Let $F_\xi = \frac{1}{m}\sum_{i=1}^m [\partial_\xi b(\xi))]^T \cdot \partial_\xi b(\xi))$. For linear baseline estimators $b(\xi)$, we have
\begin{eqnarray}
\sqrt{m} (\hat{\xi} - \xi_0) = \frac{1}{\sqrt{m}} F^{-1}_{\xi} \cdot \sum_{i=1}^{m}  \left[G(\omega_i) - b(\xi_0) \right]  \cdot \partial_\xi b(\xi_0) + O(1) 
\end{eqnarray}
\end{lemma}
\textbf{Proof Sketch.} Since $\hat{\xi}$ is the minimum mean-square-error estimator optimizing $\sum_{i=1}^m  \left[G(\omega_i) - b(\xi) \right]^2$, we have $\partial_\xi \sum_{i=1}^m  \left[G(\omega_i) - b(\xi) \right]^2=0$. Expanding the left-hand side from $\xi = {\xi}_0$ toward $\xi = \hat{\xi}$, we have $0=\partial_\xi \sum_{i=1}^m  \left[G(\omega_i) - b(\xi_0) \right]^2 + \partial^2_\xi \sum_{i=1}^m  \left[G(\omega_i) - b(\xi) \right]^2(\hat{\xi}-\xi_0)+o(||\hat{\xi} -\xi_0 ||_2$. It yields the desired result using the fact that $b(\xi)$ is linear (thus $\partial^2_\xi b(\xi) = 0$) and using the definition of $F_\xi$.





\begin{theorem} The asymptotic variance of $Z_{\rm DPE}$, using optimal estimators $\hat{\psi}$, $\hat{\eta}$, and $\hat{\xi}$, is always less than that of $Z_{\rm OIS}$ using some ${\psi}_0$, ${\eta}_0$ and ${\xi}_0$, i.e.,
\begin{eqnarray}
\text{var}(Z_{\rm DPE}) = \text{var}(Z_{\rm OIS}) - \text{var}(V_A) - \text{var}(V_B)
\end{eqnarray}
where $V_A$ and $V_B$ are projections of $\{\mu_i = {f(\omega_i; \hat{\psi})[ G(\omega_i)- b(\hat{\xi})]}/{P(\omega_{i};\hat{\eta})}, \ \forall i\}$ onto the row space of $S_\eta = \partial_\eta \log P(\omega_{i};\eta_0)$ and $S_\xi = \partial_\xi b(\xi_0)$, respectively.


\end{theorem} 

\textbf{Proof Sketch.} We provide a sketch of the proof below and include the full proof in the appendix. 

Step 1: Define auxiliary function $\mu_i=\mu(\omega_i; \eta,\psi,\xi) = \frac{f(\omega_i)[ G(\omega_i)- b(\xi_0 )]}{P(\omega_i; \eta)}$, such that $Z_{\rm DPE}$ (which is $\hat{\theta}$ in the notes with $Z_{\rm OIS}$ being $\theta$) can be written in $\sum_{i=1}^{n}\mu(x_{i};\theta,\xi,\eta) - \theta = 0$. Then expand it from $\eta_0,\psi_0,\xi_0$ to $\hat{\eta},\hat{\psi},\hat{\xi}$, to obtain $\sqrt{n} (\hat{\theta} -\theta)  = \frac{1}{\sqrt{n}} \sum_{i=1}^{n} \mu(\omega_{i};\theta,\xi_0,\eta_0) + E(\partial_\eta \mu) (\hat{\eta} - \eta) + E(\partial_\xi \mu) \sqrt{n} (\hat{\xi} - \xi) + O(1)$.
    
Step 2: Rearranging the terms, plugging in Lemma 1 and Lemma 2, and using the fact of $ \sum_{i=1}^{n}{S_\eta}'F^{-1}_{\eta} S_\eta =1$ and $
\sum_{i=1}^{n} w_i^2 {S_\xi}' F^{-1}_{\xi} S_{\xi} = 1$, we obtain the equation below, where define $S_\xi$ and $S_eta$ here. Note that we use weights $w_i=1$ throughout the proof. 

Step 3, Recognize that $S_\xi$ and $S_\eta$ are orthogonal. The two terms in C (i.e., A and B) can be viewed as projecting $\mu_i$ onto orthogonal row spaces of $S_\xi$ and $S_\eta$, respectively. Define these as $V_A$ and $V_B$  The first term on the right hand side in  $$ \sqrt{n} (\hat{\theta} -\theta) = \frac{1}{\sqrt{n}} \sum_{i=1}^{n} \{ \mu_i - \underbrace{E(\mu_i {S_\eta}')F^{-1}_{\eta} \cdot  S_\eta}_{V_A}  -  \underbrace{E(\mu_i {S_\xi}')\cdot w_i^2 F^{-1}_{\xi} \}}_{V_B} + O(1)$$ is indeed OIS since $\sqrt{n} (\hat{\theta} -\theta) = \frac{1}{\sqrt{n}} \sum_{i=1}^{n} \mu_i$. 

From Pythagorean relationship, we prove $\text{var}(Z_{\rm DPE}) = \text{var}(Z_{\rm OIS}) - \text{var}(V_A) - \text{var}(V_B)$. The theorem shows that the use of the DPE estimator always reduces the asymptotic variance of the estimator of OIS.

%% file: 5Experiments.tex
\section{Experiments}

In this section, we present an empirical study of applying Double Policy Estimator on Decision Transformer 
to verify the feasibility and effectiveness of our proposed method. 
We evaluate the performance of our proposed algorithm on the continuous control tasks from the D4RL benchmark and compare it with several popular SOTA baselines. Furthermore, we analyze some critical properties to confirm the rationality of our motivation. 


\subsection{Experiment Setup}
We empirically evaluate the performance of our proposed algorithm on the \textbf{Gym Locomotion} v2: a series of continuous control tasks consisting of \texttt{HalfCheetah, Hopper, and Walker2d } datasets from the D4RL offline reinforcement learning benchmark \cite{fu2020d4rl} with medium, medium-replay, and medium-expert datasets which include mixed and suboptimal trajectories.
Specifically, \texttt{Medium} dataset includes 1 million timesteps generated by a “medium” policy that achieves approximately one-third of the score of an expert policy; \texttt{Medium-Replay} includes 25k-400k timesteps that are gathered from the replay buffer of an agent trained to the performance of a medium policy; \texttt{Medium-Expert} includes 1 million timesteps generated by the medium policy and then concatenated with 1 million timesteps generated by an expert policy.




\subsection{Baseline Selection}
We compare our proposed algorithm to the following SOTA methods, where they aim to tackle the current challenges in offline reinforcement learning from different perspectives: Decision Transformer (DT) 
 \cite{chen2021decision}, reward-conditioned behavioral cloning (RvS) \cite{rvs}, Conservative Q-Learning (CQL) \cite{cql}, BEAR \cite{bear}, UWAC \cite{uwac}, behavior cloning (BC), and  Implicit Q Learning (IQL)\cite{iql}. CQL and IQL represent the state-of-the-art in model-free offline RL; RvS and DT represent the state-of-the-art in sequence-modeling-based supervised learning.

\subsection{DPE weights implementation}
Note when proposing double policy estimation, there is no specific limitation on how $\pi_b$ and $\pi_t$ are estimated and how DPE weights are calculated. In this empirical section, we consider the following as one possible implementation: (1) We first apply CQL to train a neural network that generates mean and variances for Gaussian distributions as maximum likelihood estimation to obtain the estimated behavior policy $\hat{\pi}_b$. (2) Then for each trajectory $\omega_i$ we can calculate the estimated behavior weights as
$\hat{w}^{\pi_b}_i = \hat{\pi}_b(a_i | \omega_i)$ (3) Next we train DT using $l_2$ loss for updating each timestep, but we record the MSE $(a_i - \hat{a_i})^2$ as the variance, and $\hat{a_i}$ as the mean for the Gaussian distribution, i.e. $\mathcal{N}({a_i}, (a_i - \hat{a_i})^2)$ as target policy estimation. (4) There are multiple ways to calculate these target weights, e.g. cumulative distribution function (CDF): $P({a_i}-\beta < \hat{a_i} \leq {a_i}+\beta) $ where $\beta$ is a probability offset, or probability density function (PDF). In this empirical result, we consider using exponentiated clipped log-likelihood: $\operatorname{exp}({l_{{a}}(\hat{a}, (a_i - \hat{a_i})^2)})$ with $l_{\hat{a}}$ clipped at 0.05 and 0.995.
\subsection{General Performance}

\input{0table}

We first evaluate and compare the performance of the proposed method with all selected baselines in terms of average reward in Table 1, where 0 represents a random policy and 100 represents an expert policy, with reward normalized per \cite{fu2020d4rl}. All results are averaged over 3 different seeds over the final 10 evaluations, we put the full results including the error bar of all baselines in the appendix. Overall, we find DPE applied DT achieves better performance than the original decision transformer on almost all datasets, and outperforms the state-of-the-art baselines over several datasets. Especially, in `medium-replay' datasets that include mixed optimal and sub-optimal trajectories, our method could bring a significant advancement in terms of reward. The finding that our proposed method attains competitive results stands in contrast to Decision Transformer which emphasizes the direct improvements brought by applying double policy estimation. 

\begin{figure*}[h]
     \centering
     \begin{subfigure}[b]{0.325\textwidth}
         \centering
         \includegraphics[width=\textwidth]{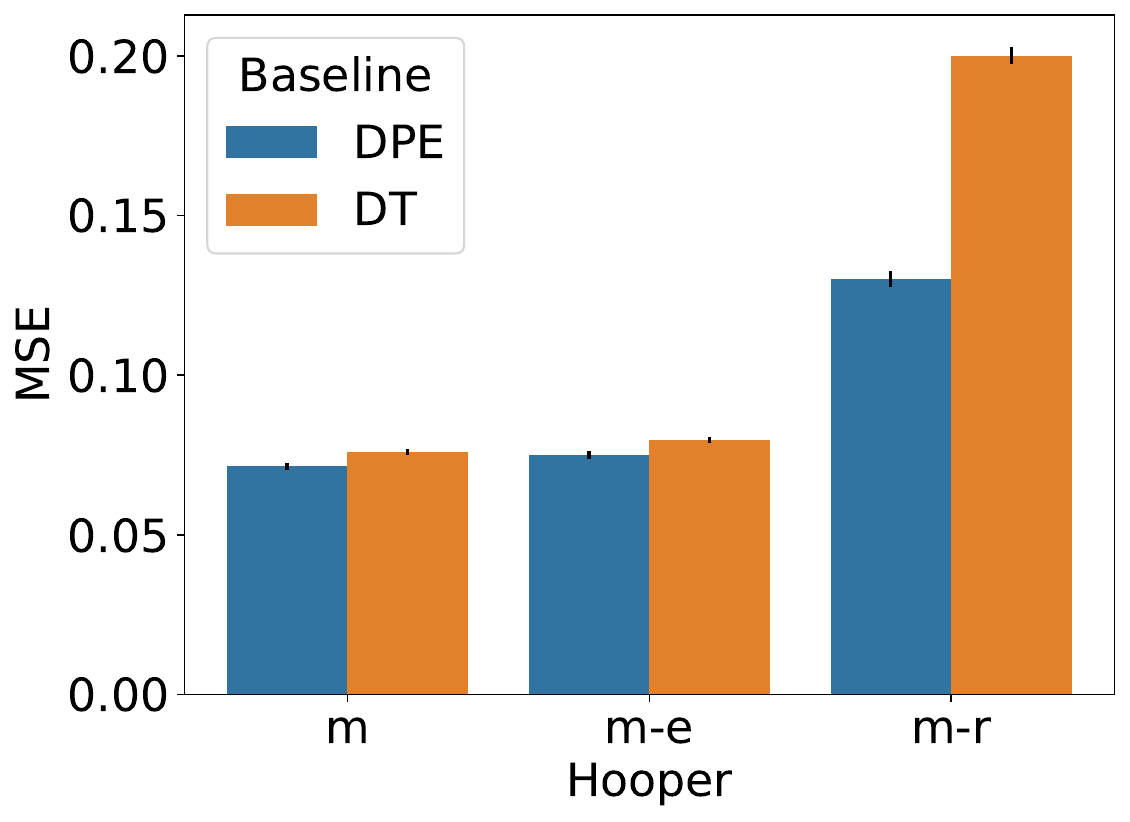}
         \caption{Hopper-medium}
         \label{fig:one}
     \end{subfigure}
     \hfill
     \begin{subfigure}[b]{0.325\textwidth}
         \centering
         \includegraphics[width=\textwidth]{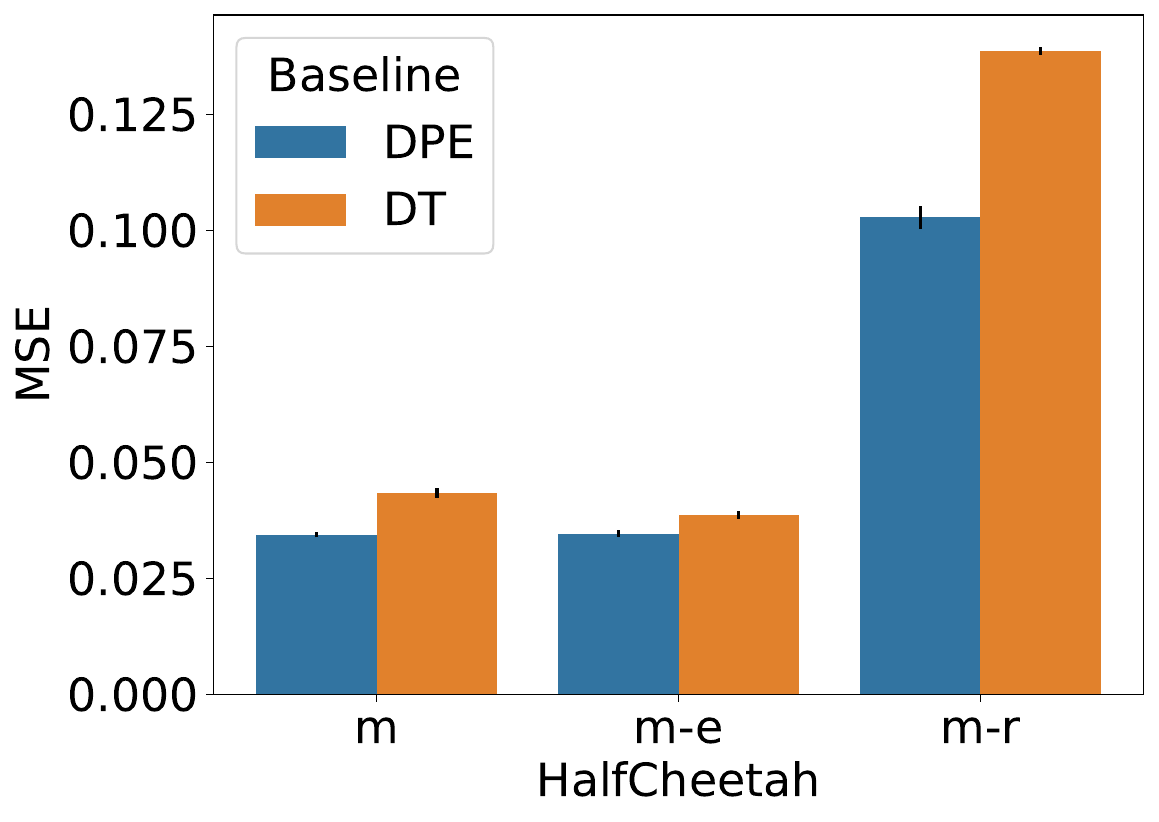}
        \caption{HalfCheetah-medium}
         \label{fig: two}
     \end{subfigure}
     \hfill
     \begin{subfigure}[b]{0.325\textwidth}
         \centering
         \includegraphics[width=\textwidth]{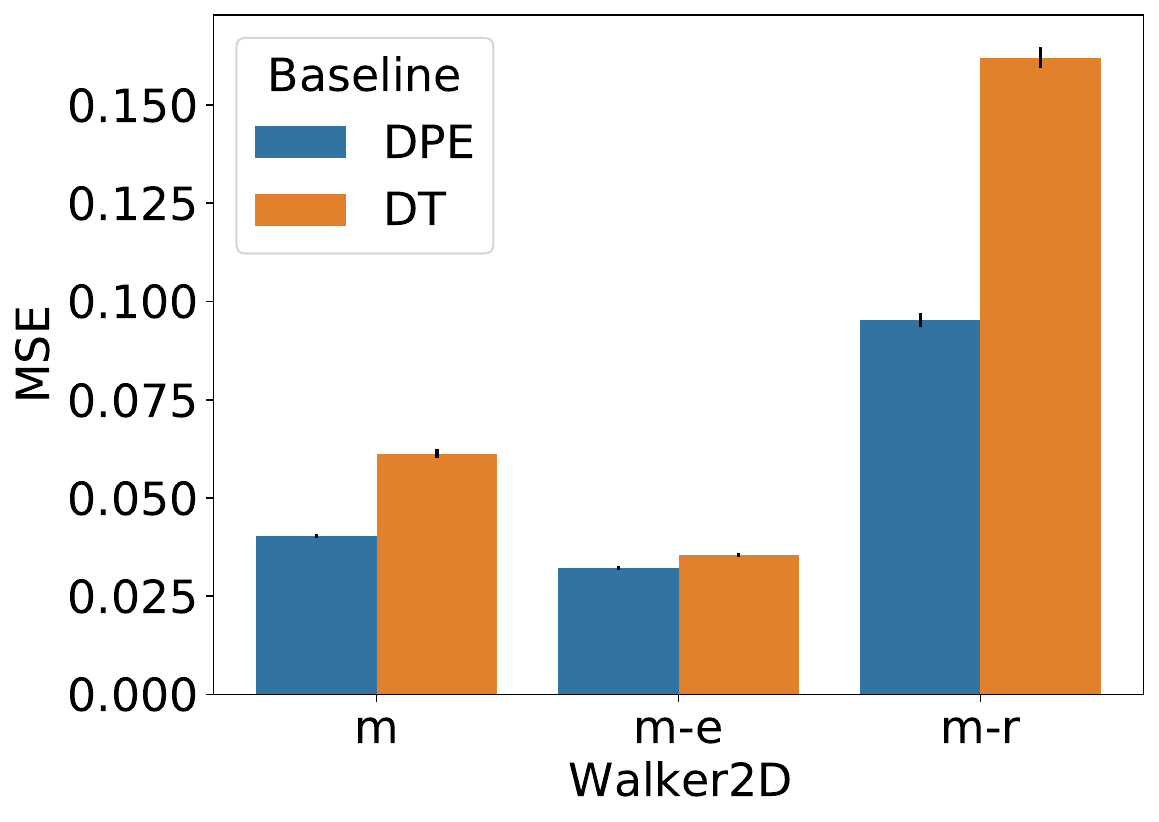}
        \caption{Walker2D-medium}
         \label{fig:tres}
     \end{subfigure}

    \caption{MSE comparison with DT and DPE}
\end{figure*}

\begin{figure*}[th!]
     \centering
     \begin{subfigure}[b]{0.325\textwidth}
         \centering
         \includegraphics[width=\textwidth]{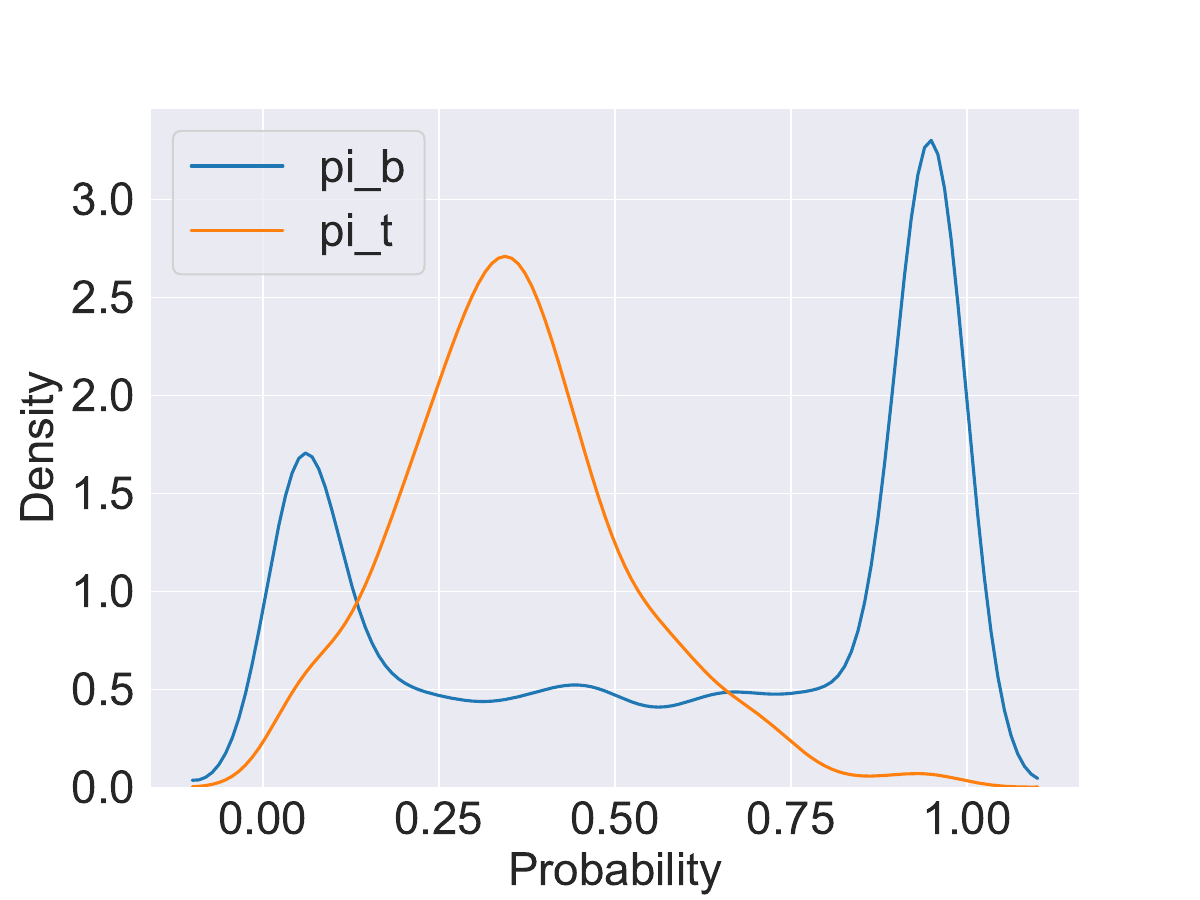}
         \caption{Hopper-medium}
         \label{fig:one}
     \end{subfigure}
     \hfill
     \begin{subfigure}[b]{0.325\textwidth}
         \centering
         \includegraphics[width=\textwidth]{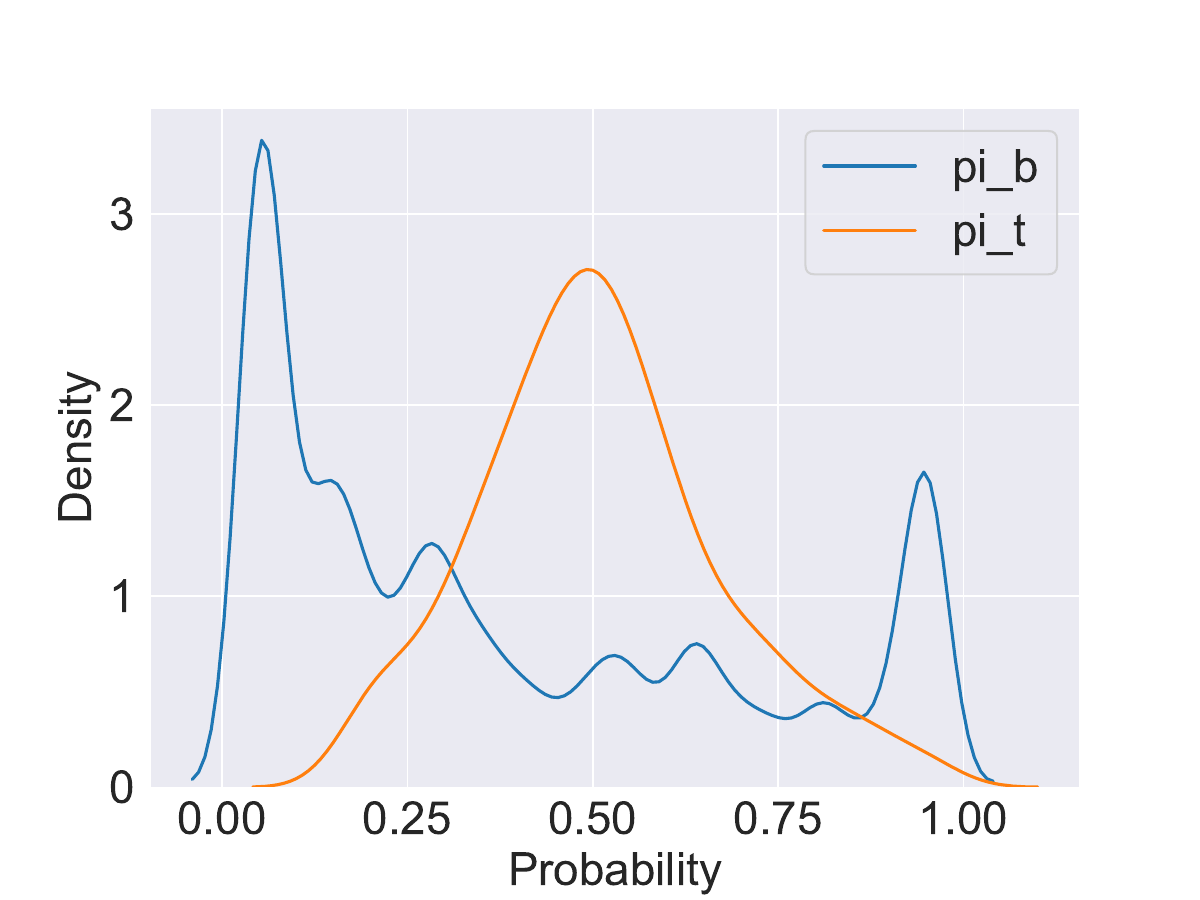}
        \caption{Hopper-medium-replay}
         \label{fig: two}
     \end{subfigure}
     \hfill
     \begin{subfigure}[b]{0.325\textwidth}
         \centering
         \includegraphics[width=\textwidth]{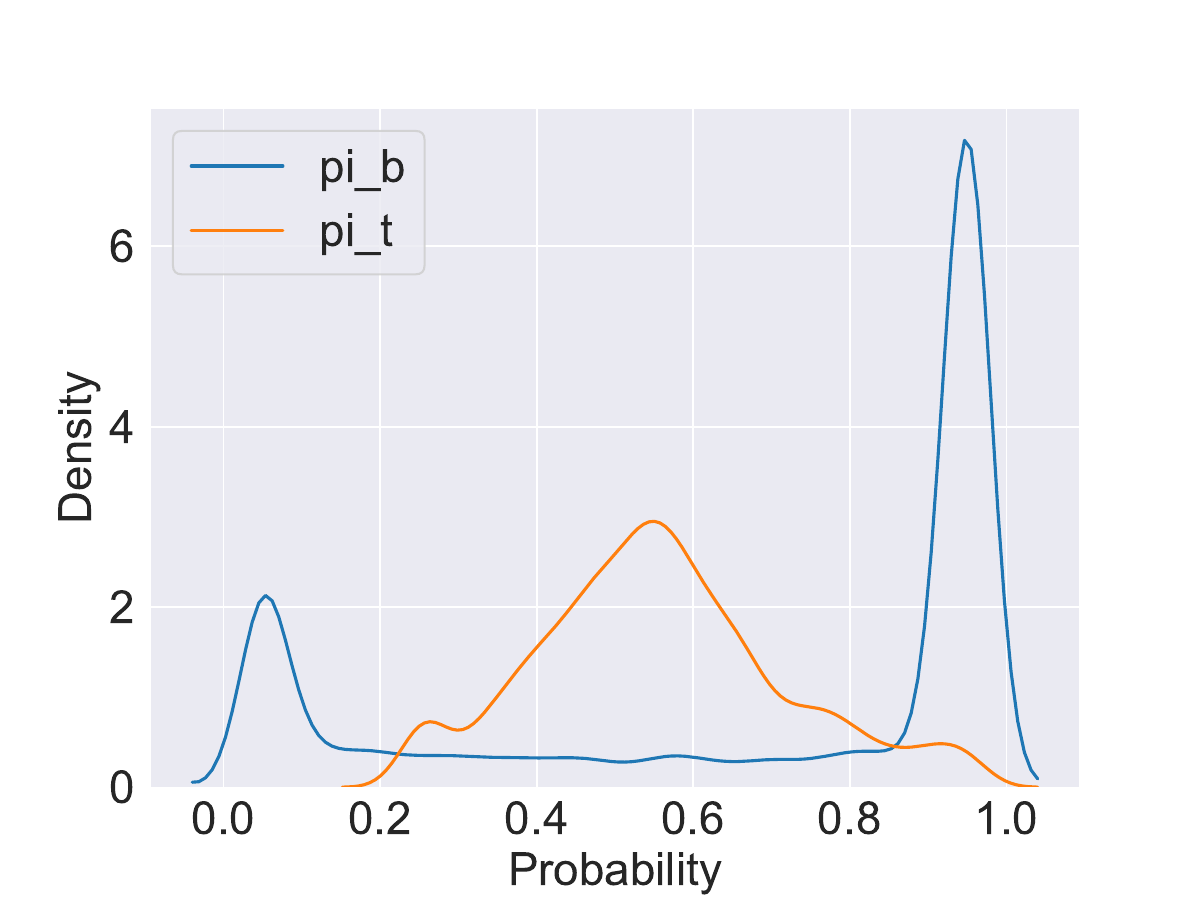}
        \caption{Hopper-medium-expert}
         \label{fig:tres}
     \end{subfigure}

    \caption{Comparing Kernel Density Estimate of estimated $\pi_b$ and $\pi_t$ on \texttt{Hopper} datasets.}
\end{figure*}

\subsection{Discussions}
To demonstrate the actual effectiveness of reducing the variance, we also record the MSE from the final evaluation stage of both DPE and DT for off-policy evaluation in Fig. 2, the results show that using DPE weights could bring a generally lower MSE on all environments selected compared to DT, validating our efficiency on variance reduction. To visualize the source of effectiveness in the double importance weights estimation we record the distribution of $\pi_b$ and $\pi_t$ on the `hopper' environment and provide a kernel density estimate plot in Fig. 3. The drastic difference from the two distributions could mean that the behavior policy estimated are acting as a correction weight to offset the probability sampling from the target policy distribution, leading to improved performance and reduced variance. As an example, an occasional sub-optimal trajectory that the target trajectory learned with high probability could be corrected by the low probability from the estimated behavior policy, making this a low-weight trajectory to learn from.


\begin{wrapfigure}{r}{.4\textwidth}
\vspace{-.4in}
\centering
\includegraphics[width=0.4\textwidth]{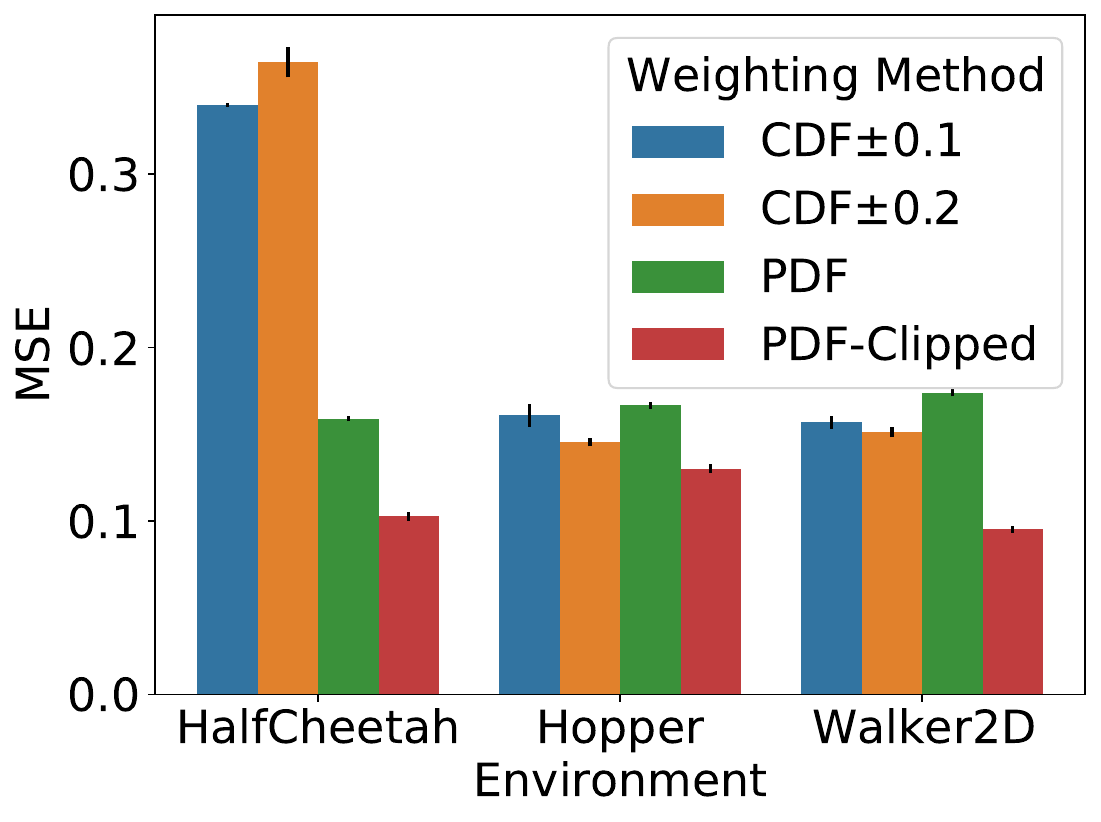}
\caption{Ablations results on comparing different probability sampling methods on estimated $\pi_b$}
\end{wrapfigure}


\subsection{Ablation Studies}
According to the object of DPE, the estimation of $\pi_b$ still determines the target policy weights. In this section, we evaluate and compare several different ways to calculate the exact probability generated from the estimated behavior distribution marking as CDF $\pm$0.1, CDF $\pm$0.2, PDF, clipped PDF,
and demonstrate the results over \texttt{medium-replay} datasets in terms of MSE in Figure 3. We see that despite some cases, most of the settings are similar regarding their prospective MSE, indicating that when a proper estimation of this Gaussian distribution is obtained, their method of sampling probability is not a major concern. Nevertheless, we find that using a clipped PDF for behavior probability selection brings the lowest MSE in general.

%% file: 0table.tex

\begin{table}[ht]
\centering
\resizebox{\textwidth}{!}{%
\begin{tabular}{@{}cccccccccc@{}}
\toprule
Dataset    & Environment    & DPE   & DT    & RvS   & CQL   & BEAR & UWAC & BC   & IQL   \\ \midrule
\multirow{3}{*}{medium} & HalfCheetah & 45.4$\pm$0.3 & 42.6$\pm$0.1 & 41.6 & 44.4 & 41.7 & 42.2 & 43.1 & \textbf{47.4} \\
           & Hopper         & \textbf{69.8$\pm$1.9}  & 67.6$\pm$1.0  & 60.2  & 58.8  & 52.1 & 50.9 & 63.9 & 66.3  \\
           & Walker         & 77.9$\pm$0.8  & 74.0$\pm$1.4  & 71.7  & \textbf{79.2}  & 59.1 & 75.4 & 77.3 & 78.3  \\ \midrule
\multirow{3}{*}{\begin{tabular}[c]{@{}c@{}}medium-\\ replay\end{tabular}} & HalfCheetah & 40.5$\pm$1.5         & 36.6$\pm$0.8         & 38.0 & \textbf{46.2} & 38.6 & 35.9 & 4.3  & 44.2 \\
           & Hopper         & \textbf{94.6$\pm$0.7}  & 79.4$\pm$7.0  & 73.5  & 48.6  & 33.7 & 25.3 & 30.9 & 94.5  \\
           & Walker         & \textbf{83.5$\pm$1.2 } & 66.6$\pm$3.0  & 60.6  & 26.7  & 19.2 & 23.6 & 36.9 & 73.9  \\ \midrule
\multirow{3}{*}{\begin{tabular}[c]{@{}c@{}}medium-\\ expert\end{tabular}} & HalfCheetah & 82.5$\pm$5.8         & \textbf{87.8$\pm$2.6 }        & 92.2 & 62.4 & 53.4 & 42.7 & 59.9 & 86.7 \\
           & Hopper         & \textbf{108.2$\pm$1.6} & 107.6$\pm$1.8 & 101.7 & 104.6 & 96.3 & 44.9 & 79.6 & 91.5  \\
           & Walker         & 93.7$\pm$6.2  & 108.1$\pm$0.2 & 106.0 & 108.1 & 40.1 & 96.5 & 36.6 & \textbf{109.6} \\ \midrule
\multicolumn{2}{c}{average}           & \textbf{77.34}  & 74.60         & 71.72 & 64.33 & 48.24 & 48.60 & 48.06 & 76.93 \\ \hline
\end{tabular}%
}

\caption{Overall performance of the normalized score of selected baselines on D4RL benchmark. All results are evaluated on 'v2' environments and datasets.}
\label{tab:my-table}
\end{table}

%% file: 6Conclusion.tex
\section{Limitations and Social Impact}
There are several opportunities for future work. First, our approach requires a warm-up phase prior to the training of the decision transformer to obtain the estimated behavior policy. Also, as RvS methods perform poorly in stochastic environments as pointed out in \cite{paster2022you}, the currently proposed method cannot resolve such issues. 
We believe this work will result in positive social impacts as this will help to avoid unexpected behaviors from occasional unwanted trajectories from the dataset and make the trained model more stable. However, this can potentially be used for making harmful decisions under training with specific harmful and biased datasets.

\section{Conclusion}
In this paper, we present DPE, a double policy estimation for importance sampling methods that are proven statistically efficient for variance reduction for Off-Policy Evaluation in Sequence-Modeled Reinforcement Learning. 
Computing both the behavior policy estimate and target estimate from the same set of
data allows DPE to correct for the sampling error inherent to importance sampling with the true behavior policy in the offline dataset. We evaluated DPE applied decision transformer across several benchmarks against current and SOTA works and showed that it demonstrated competitive performances while improving the evaluation results of the Decision Transformer, especially on the dataset filled with sub-optimal trajectories, and confirming the effect of variance reduction through MSE comparison. Finally, we studied the possible cause for such improvements by visualizing the density of the estimated target policy and behavior policy. 


%% file: 7appendix.tex
\appendix


\section{Mathematical Details}
\subsection{Notations and Explanations}
We use Table 1 to summarize the notations introduced in this paper and their corresponding explanations, more explanations can be found when first introduced in the main paper.

\begin{table}[h]
\centering
\resizebox{0.65\textwidth}{!}{%
\begin{tabular}{cc}
\hline
Notation                         & Definition                           \\ \hline
s                                & state space                          \\
a                                & action space                         \\
P                                & environment transition probabilities \\
r                                & reward, reward function              \\
$\mathcal{D}$                    & offline dataset                      \\
$\omega$                         & a trajectory                         \\
H                                & Horizon length                       \\
$\pi_e$                          & evaluation policy                    \\
$\pi_b$                          & behaviour policy                     \\
$\pi_t$                          & target policy                        \\
$\nu$                             & density ratio                        \\
Z                                & importance weighed policy gradient   \\
$g$                              & policy score                         \\
q                                & return-to-go                         \\
$\hat{\pi}_{b}$                  & estimated behavior policy            \\
$\hat{\pi}_{t}$                  & estimated target policy              \\
$b^{\eta}$                       & baseline predicted return-to-go      \\
$\mathcal{N}(a, \sigma^2)$ & Gaussian distribution, with mean of a and std of $\sigma$ \\
Var                              & Variance Operator                    \\
$\hat{\xi},\hat{\eta},\hat{\psi}$ & estimators                           \\ \hline
\end{tabular}%
}
\caption{}
\label{tab:my-table}
\end{table}

\subsection{DPE for sequence modeling-based reinforcement learning}
Let $\mathcal{D}$ be a set of off-policy trajectories of length $H+1$ collected by a behavior policy $\pi_b$, denoted by $\mathcal{D}=\{\omega_i, \ \forall i \}$ with each trajectory ${\omega}_i= \{({s_0}^{(i)}, {a_0}^{(i)}, {r_0}^{(i)}, \cdots, {s_H}^{(i)}, {a_H}^{(i)}, {r_H}^{(i)})$).

We first denote the approximated behavior policy with a Gaussian distribution as $\pi_b = {argmax}_{\pi} P(\pi | \omega)$, and the approximated target policy as $\pi_t = \hat{a_t} + \hat{\sigma^2}*n_k$, where $\hat{a_t}$ and $\hat{\sigma^2}$is the mean and variance generated from the decision transformer,  $n_k$ is Gaussian noise.

\subsection{Proof For Lemma 1}

\begin{lemma} Let $F_\eta = -\frac{1}{m}\sum_{i=1}^m \partial^2_\eta \log P(\omega_{i};\hat{\eta_0}) $ be the Fisher Information Matrix. We have
\begin{eqnarray}
\sqrt{m} (\hat{\eta} - \eta_0) = \frac{1}{\sqrt{m}} F^{-1}_{\eta} \cdot \sum_{i=1}^{m} \partial_\eta log P(\omega_{i};\eta_0)  + O(1) 
\end{eqnarray}
\end{lemma}

\textit{Proof.}
Since $\hat{\eta}$ is the maximum likelihood estimator that optimizes $P(\omega_{i};{\eta})$, we have $\partial_\eta \sum_{i=1}^m \log P(\omega_{i};{\eta}) =0 $ at $\eta = \hat{\eta}$: 

\begin{eqnarray}
0 && = \sum_{i=1}^{n}  \partial_\eta log P(\omega_{i};\eta_0) +  \sum_{i=1}^{n}  \partial_\eta^2 log P(\omega_{i};\hat{\eta}_0)(\hat{\eta} - \eta_0) + O(||\hat{\eta} - \eta||_2) \nonumber \\
 && \text{Expanding the right-hand side from $\eta = {\eta}_0$ toward $\eta = \hat{\eta}$, we have}  \nonumber \\
\sqrt{n} (\hat{\eta} - \eta_0) &&= \frac{1}{\sqrt{n}} \{ -\frac{1}{n}  \partial_\eta^2 log P(\omega_{i};\eta_0)(\hat{\eta} - \eta_0) \}^{-1} \cdot \sum_{i=1}^{n} \partial_\eta log P(\omega_{i};\eta_0)  + O(1) \nonumber \\
\sqrt{n} (\hat{\eta} - \eta_0) &&= \frac{1}{\sqrt{n}} \{ -\frac{1}{n}  \partial_\eta^2 log P(\omega_{i};\eta_0)(\hat{\eta} - \eta_0) \}^{-1} \cdot \sum_{i=1}^{n} \partial_\eta log P(\omega_{i};\eta_0)  + O(1) \nonumber \\
\sqrt{n} (\hat{\eta} - \eta_0) &&= \frac{1}{\sqrt{n}} F^{-1}_{\eta} \cdot \sum_{i=1}^{n} \partial_\eta log P(\omega_{i};\eta_0)  + O(1) 
\end{eqnarray}


which yields the desired result by rearranging the terms and leveraging the Fisher Information Matrix $F_\eta$.

\subsection{Proof For Lemma 2}

\begin{lemma} Let $F_\xi = \frac{1}{m}\sum_{i=1}^m [\partial_\xi b(\xi))]^T \cdot \partial_\xi b(\xi))$. For linear baseline estimators $b(\xi)$, we have
\begin{eqnarray}
\sqrt{m} (\hat{\xi} - \xi_0) = \frac{1}{\sqrt{m}} F^{-1}_{\xi} \cdot \sum_{i=1}^{m}  \left[G(\omega_i) - b(\xi_0) \right]  \cdot \partial_\xi b(\xi_0) + O(1) 
\end{eqnarray}
\end{lemma}

\textit{Proof. Since $\hat{\xi}$ is the minimum mean-square-error estimator optimizing $\sum_{i=1}^m  \left[G(\omega_i) - b(\xi) \right]^2$, we have: }
\begin{eqnarray}
\partial_\xi \sum_{i=1}^m  \left[G(\omega_i) - b(\xi) \right]^2=0
\end{eqnarray}
Expanding the right-hand side from $\xi = {\xi}_0$ toward $\xi = \hat{\xi}$, we have
\begin{eqnarray}
    && 0  =\partial_\xi \sum_{i=1}^m  \left[G(\omega_i) - b(\xi_0) \right]^2 + \partial^2_\xi \sum_{i=1}^m  \left[G(\omega_i) - b(\xi) \right]^2(\hat{\xi}-\xi_0)+o(||\hat{\xi} -\xi_0 ||_2 \nonumber \\
    && 0 = \sum_{i=1}^{n} \partial_\xi w_i^2 [G(\omega_i)-b(\xi)]^2 + \sum_{i=1}^{n} \partial_\xi^2 w_i^2 [G(\omega_i)-b(\xi)]^2 (\hat{\xi} - \xi)^2 + O(||(\hat{\xi} - \xi)||_2) \nonumber \\
    && 0 = \sum_{i=1}^{n} (-I) \cdot [G(\omega_i)-b(\xi)]^2 w_i^2 \cdot \partial_\xi b(\xi_0) + \sum_{i=1}^{n} [w_i^2(\partial_\xi b(\xi_0))] ^2 - I[G(\omega_i)-b(\xi_0) \cdot   w_i^2 \cdot\partial_\xi^2 b(\xi_0) ] ^2 (\hat{\xi} - \xi) + O(1) \nonumber \\
    && \sqrt{n} (\hat{\xi} - \xi)  = \frac{1}{\sqrt{n}}\{\frac{1}{n} \sum_{i=1}^{n} (\partial_\xi^2 b(\xi_0))^2 w_i^2 \} ^{-1} \cdot \sum_{i=1}^{n} [G(\omega_i)-b(\xi_0)]  \cdot w_i^2 \partial_\xi b(\xi_0) + O(1) \nonumber \\
    && \sqrt{n} (\hat{\xi} - \xi) =  \frac{1}{\sqrt{n}} F^{-1}_{\xi} \cdot \sum_{i=1}^{n}  [G(\omega_i)-b(\xi_0)]  w_i^2 \partial_\xi b(\xi_0)
\end{eqnarray}
It yields the desired result using the fact that $b(\xi)$ is linear (thus $\partial^2_\xi b(\xi) = 0$) and using the definition of $F_\xi$.

\textbf{Proof Sketch.} Since $\hat{\xi}$ is the minimum mean-square-error estimator optimizing $\sum_{i=1}^m  \left[G(\omega_i) - b(\xi) \right]^2$, we have $\partial_\xi \sum_{i=1}^m  \left[G(\omega_i) - b(\xi) \right]^2=0$. Expanding the left-hand side from $\xi = {\xi}_0$ toward $\xi = \hat{\xi}$, we have $0=\partial_\xi \sum_{i=1}^m  \left[G(\omega_i) - b(\xi_0) \right]^2 + \partial^2_\xi \sum_{i=1}^m  \left[G(\omega_i) - b(\xi) \right]^2(\hat{\xi}-\xi_0)+o(||\hat{\xi} -\xi_0 ||_2$. It yields the desired result using the fact that $b(\xi)$ is linear (thus $\partial^2_\xi b(\xi) = 0$) and using the definition of $F_\xi$.

\subsection{Proof For Theorem 1}
\begin{theorem} The asymptotic variance of $Z_{\rm DPE}$, using optimal estimators $\hat{\psi}$, $\hat{\eta}$, and $\hat{\xi}$, is always less than that of $Z_{\rm OIS}$ using some ${\psi}_0$, ${\eta}_0$ and ${\xi}_0$, i.e.,
\begin{eqnarray}
\text{var}(Z_{\rm DPE}) = \text{var}(Z_{\rm OIS}) - \text{var}(V_A) - \text{var}(V_B)
\end{eqnarray}
where $V_A$ and $V_B$ are projections of $\{\mu_i = {f(\omega_i; \hat{\psi})[ G(\omega_i)- b(\hat{\xi})]}/{P(\omega_{i};\hat{\eta})}, \ \forall i\}$ onto the row space of $S_\eta = \partial_\eta log P(\omega_{i};\eta_0)$ and $S_\xi = \partial_\xi b(\xi_0)$, respectively.
\end{theorem} 

\textit{Proof.} We define auxiliary function $\mu_i=\mu(\omega_i; \eta,\psi,\xi) = \frac{f(\omega_i)[ G(\omega_i)- b(\xi_0 )]}{P(\omega_{i}; \eta)}$, such that $Z_{\rm DPE}$ (which is $\hat{\theta}$ in the notes with $Z_{\rm OIS}$ being $\theta$) can be written in $$\sum_{i=1}^{n}\mu(\omega_{i};\theta,\xi,\eta) - \theta = 0$$

Then expand it from $\eta_0,\psi_0,\xi_0$ to $\hat{\eta},\hat{\psi},\hat{\xi}$, we have

\begin{eqnarray}
0 &&= \frac{1}{n} \mu(\omega_i;\theta,\xi_0,\eta_0) - (\hat{\theta} - \theta) + \frac{1}{n} \sum_{i=1}^{n} \partial_\eta \mu(\omega_i;\theta,\xi_0,\eta_0) (\eta - \hat{\eta}) + \frac{1}{n}\sum_{i=1}^{n} \partial_\xi  \mu(\omega_i;\theta,\xi_0,\eta_0)(\xi - \hat{\xi}) \nonumber \\
&& \ \ \ \ \  + \frac{1}{n} \sum_{i=1}^{n} \partial_\xi \mu(\omega_i;\theta,\xi_0,\eta_0)(\hat{\xi} - \xi) + O(||\hat{\theta} - \theta||_2 + ||\hat{\eta} - \eta||_2 + ||\hat{\xi} - \xi||_2)\nonumber  \\
\sqrt{n} (\hat{\theta} -\theta) & &= \frac{1}{\sqrt{n}}  \sum_{i=1}^{n} \mu(\omega_i;\theta,\xi_0,\eta_0) + E(\partial_\eta \mu) (\hat{\eta} - \eta) + E(\partial_\xi \mu) \sqrt{n} (\hat{\xi} - \xi) + O(1)
\end{eqnarray}
Rearranging the terms, plugging in Lemma 1 and Lemma 2, and using the fact of $ \sum_{i=1}^{n}{S_\eta}'F^{-1}_{\eta} S_\eta =1$ and $
\sum_{i=1}^{n} w_i^2 {S_\xi}' F^{-1}_{\xi} S_{\xi} = 1$, we obtain the equation below, where define $S_\xi$ and $S_\eta$ here. Note that we use weights $w_i=1$ throughout the proof. Recognize that $S_\xi$ and $S_\eta$ are orthogonal. The two terms in $V_A$ and $V_B$ can be viewed as projecting $\mu_i$ onto orthogonal row spaces of $S_\xi$ and $S_\eta$, respectively. Define these as $V_A$ and $V_B$    

\begin{eqnarray}
\sqrt{n} (\hat{\theta} -\theta) && = \frac{1}{\sqrt{n}} \sum_{i=1}^{n} \{ \mu_i - \underbrace{E(\mu_i {S_\eta}')F^{-1}_{\eta} \cdot  S_\eta}_{V_A}  -  \underbrace{E(\mu_i {S_\xi}')\cdot w_i^2 F^{-1}_{\xi} \}}_{V_B} + O(1) \nonumber \\
\sqrt{n} (\hat{\theta} -\theta) && =  \frac{1}{\sqrt{n}} \sum_{i=1}^{n} \mu_i + E\{ - \frac{f(\omega_i) [G(\omega_i)-b(\xi_0)]  }{P(\omega_i;\eta_0) } \cdot \frac{\partial_\eta P(\omega_i;\eta_0) }{P(\omega_i;\eta_0)} \cdot \sqrt{n} (\hat{\eta} - \eta_0) \} \nonumber \\
 && \ \ \ \ \ +  E\{- \frac{f(\omega_i)}{P(\omega_i;\eta_0) } \partial_\xi b(\xi_0) \} \cdot \sqrt{n} (\hat{\xi} - \xi) + O(1) \nonumber \\
\sqrt{n} (\hat{\theta} -\theta) && =  \frac{1}{\sqrt{n}} \sum_{i=1}^{n} \mu_i - E(\mu_i \cdot  {s_\eta}') \cdot \frac{1}{\sqrt{n}} F^{-1}_{\eta} \cdot \sum_{i=1}^{n} s_\eta \nonumber \\
 && \ \ \ \ \  -  E\{- \frac{f(\omega_i)}{P(\omega_i; \eta_0) } \partial_\xi b(\xi_0) \} \cdot \frac{1}{\sqrt{n}} F^{-1}_{\xi}  \sum_{i=1}^{n}  w_i^2  [G(\omega_i)-b(\xi_0)]  \partial_\xi b(\xi_0) + O(1) \nonumber \\
= && \frac{1}{\sqrt{n}} \sum_{i=1}^{n} \mu_i -\frac{1}{\sqrt{n}}   E(\mu_i \cdot  {S_\eta}') \cdot F^{-1}_{\eta} \cdot  S_\eta
- \frac{1}{\sqrt{n}} \sum_{i=1}^{n} E(\mu_i {S_\eta}' ) \cdot  w_i^2 F^{-1}_{\eta} \cdot {S_\xi} + O(1 + \frac{2}{\sqrt{n}}) \nonumber \\
\end{eqnarray}
\text{Recall that}
$$\left\{\begin{matrix}
 \sum_{i=1}^{n}{S_\eta}'F^{-1}_{\eta} S_\eta =1 ,  
\\
 \\
\sum_{i=1}^{n} w_i^2 {S_\xi}' F^{-1}_{\xi} S_{\xi} = 1
\\
\end{matrix}\right. $$
We have:
\begin{eqnarray}
\sqrt{n} (\hat{\theta} -\theta) = \frac{1}{\sqrt{n}} \sum_{i=1}^{n} \{ \mu_i - \underbrace{E(\mu_i {S_\eta}')F^{-1}_{\eta} \cdot  S_\eta}_{V_A}  -  \underbrace{E(\mu_i {S_\xi}')\cdot w_i^2 F^{-1}_{\xi} \}}_{V_B} + O(1) \\
\end{eqnarray}

We note that ${S_\xi}'$ and ${S_\eta}'$ are orthogonal, i.e.,  

$$
\sum_{i=1}^{n}{S_\eta} {S_\xi} ={S_\xi} \sum_{i=1}^{n} {S_\eta} = 0
$$

Recognize that $S_\xi$ and $S_\eta$ are orthogonal. The two terms in Eq.(10) can be viewed as projecting $\mu_i$ onto orthogonal row spaces of $S_\xi$ and $S_\eta$, respectively. Define these as $V_A$ and $V_B$  The first term on the right hand side in  $$ \sqrt{n} (\hat{\theta} -\theta) = \frac{1}{\sqrt{n}} \sum_{i=1}^{n} \{ \mu_i - \underbrace{E(\mu_i {S_\eta}')F^{-1}_{\eta} \cdot  S_\eta}_{V_A}  -  \underbrace{E(\mu_i {S_\xi}')\cdot w_i^2 F^{-1}_{\xi} \}}_{V_B} + O(1)$$ is indeed OIS since $\sqrt{n} (\hat{\theta} -\theta) = \frac{1}{\sqrt{n}} \sum_{i=1}^{n} \mu_i$. From Pythagorean relationship, we prove $\text{var}(Z_{\rm DPE}) = \text{var}(Z_{\rm OIS}) - \text{var}(V_A) - \text{var}(V_B)$. The theorem shows that the use of the DPE estimator always reduces the asymptotic variance of the estimator of OIS. 

From Pythagorean relationship, we prove $\text{var}(Z_{\rm DPE}) = \text{var}(Z_{\rm OIS}) - \text{var}(V_A) - \text{var}(V_B)$. The theorem shows that the use of the DPE estimator always reduces the asymptotic variance of the estimator of OIS.

\begin{algorithm}[h]
    \caption{pseudocode for DPE}
    \begin{algorithmic}[1]
        \State {Initiate $\hat{\theta}$ for $\pi_{\theta}(a|s)$}
        \For{$k = 0 $ to $pretrain\_steps$}                    
            \State {Random Sample Trajectories: $\boldsymbol{\tau} \sim \mathrm{D}$}
            \State {Sample time index for each trajectory: $h \sim {\tau}_{i}[1,L]$}
            \State {Calculate Loss: $\mathcal{L}(\hat{\theta}) = \sum_{s_t, a_t, h}\text{log} \pi_{\theta}(a_t|s_t)$}
            \State {Update policy parameters: $\hat{\theta} = \hat{\theta} + \eta \nabla_{\theta}(\mathcal{L}(\hat{theta}))$}
        \EndFor   
        \State {initialize $\hat{\pi_t}$, Decision Transformer}
        \For{$k = 0 $ to $max\_train\_steps$}
            \State {Random Sample Trajectories: $\boldsymbol{\tau} \sim \mathrm{D}$}
            \State {Sample time index for each trajectory: $h \sim {\tau}_{i}[1,L]$}
            \State {Generate Trajectories from DT: $\hat{a} = \text{DT}(R, s, a, t)$}
            \State {Calculate Loss for DT: $\mathcal{L}(\Tilde{\theta}) = \frac{1}{N}\sum_{i=1}^{N}(\hat{a_i}-a_i)^2$}
            \State {Estimate $\nabla_{\theta} J(\pi_{t}) = \sum_{i,t} \nabla_{\theta_t} \text{log}\pi_t(a_{i,t}|s_{it})A((a_{i,t}|s_{it}))$}
            \State {Update Decision Transformer: $\theta_{\pi_t} = \theta_{\pi_t}  + \eta \nabla_{\theta_t}J(\pi_{\theta_t})$ }
        \EndFor
        \State {Return Decision Transformer $\pi_t$}
    \end{algorithmic}
\end{algorithm}

\section{Experimental Details}
Code for experiments can be found on GitHub.
\subsection{Implementation Details}
Our code is based on the original Decision Transformer\cite{chen2021decision} and CQL\cite{cql}. We summarize the pseudocode for DPE training process in Algorithm 1. The hyperparameters used are shown below.

\begin{table}[ht]
\caption{Hyperparameters of DPE in experiments for D4RL Dataset.}
\vskip 0.15in
\begin{center}
\begin{small}
\begin{tabular}{ll}
\toprule
\textbf{Hyperparameter} & \textbf{Value}  \\
\midrule
Number of layers & $3$  \\ 
Number of attention heads    & $1$  \\
Embedding dimension    & $128$  \\
Nonlinearity function & ReLU \\
Batch size   & $64$ \\
Context length $K$ & $20$ HalfCheetah, Hopper, Walker \\
Return-to-go conditioning   & $6000$ for HalfCheetah \\
                            & $3600$ for Hopper \\
                            & $5000$ for Walker \\
Dropout & $0.1$ \\
Learning rate & $10^{-4}$ \\
Grad norm clip & $0.25$ \\
Weight decay & $10^{-4}$ \\
Learning rate decay & Linear warmup for first $10^5$ training steps \\
\bottomrule
\end{tabular}
\end{small}
\end{center}
\end{table}

%% file: main.bbl
\begin{thebibliography}{64}
\providecommand{\natexlab}[1]{#1}
\providecommand{\url}[1]{\texttt{#1}}
\expandafter\ifx\csname urlstyle\endcsname\relax
  \providecommand{\doi}[1]{doi: #1}\else
  \providecommand{\doi}{doi: \begingroup \urlstyle{rm}\Url}\fi

\bibitem[Luo et~al.(2022)Luo, Ma, Munden, Wu, and Jiang]{luo2022multisource}
Xiaoling Luo, Xiaobo Ma, Matthew Munden, Yao-Jan Wu, and Yangsheng Jiang.
\newblock A multisource data approach for estimating vehicle queue length at
  metered on-ramps.
\newblock \emph{Journal of Transportation Engineering, Part A: Systems},
  148\penalty0 (2):\penalty0 04021117, 2022.

\bibitem[He et~al.(2022)He, Wang, Han, Zou, and Miao]{He2022Robust}
Sihong He, Yue Wang, Shuo Han, Shaofeng Zou, and Fei Miao.
\newblock A robust and constrained multi-agent reinforcement learning framework
  for electric vehicle amod systems.
\newblock \emph{arXiv preprint arXiv:2209.08230}, 2022.

\bibitem[Hua et~al.(2023)Hua, Zhang, Zhang, Li, Yu, Xu, and
  Zhou]{HUA2023121526}
Min Hua, Cetengfei Zhang, Fanggang Zhang, Zhi Li, Xiaoli Yu, Hongming Xu, and
  Quan Zhou.
\newblock Energy management of multi-mode plug-in hybrid electric vehicle using
  multi-agent deep reinforcement learning.
\newblock \emph{Applied Energy}, 348:\penalty0 121526, 2023.

\bibitem[Ma et~al.(2020)Ma, Karimpour, and Wu]{ma2020statistical}
Xiaobo Ma, Abolfazl Karimpour, and Yao-Jan Wu.
\newblock Statistical evaluation of data requirement for ramp metering
  performance assessment.
\newblock \emph{Transportation Research Part A: Policy and Practice},
  141:\penalty0 248--261, 2020.

\bibitem[Chen and Lan(2023)]{chen2023minimizing}
Jingdi Chen and Tian Lan.
\newblock Minimizing return gaps with discrete communications in decentralized
  pomdp, 2023.

\bibitem[Gogineni et~al.(2023)Gogineni, Mei, Wei, Lan, and
  Venkataramani]{gogineni2023accmer}
Kailash Gogineni, Yongsheng Mei, Peng Wei, Tian Lan, and Guru Venkataramani.
\newblock Accmer: Accelerating multi-agent experience replay with cache
  locality-aware prioritization.
\newblock \emph{arXiv preprint arXiv:2306.00187}, 2023.

\bibitem[Sutton and Barto(2018)]{sutton2018reinforcement}
Richard~S Sutton and Andrew~G Barto.
\newblock \emph{Reinforcement learning: An introduction}.
\newblock MIT press, 2018.

\bibitem[Chen et~al.(2021{\natexlab{a}})Chen, Umrawal, Lan, and
  Aggarwal]{chen2021deepfreight}
Jiayu Chen, Abhishek~K Umrawal, Tian Lan, and Vaneet Aggarwal.
\newblock Deepfreight: A model-free deep-reinforcement-learning-based algorithm
  for multi-transfer freight delivery.
\newblock In \emph{Proceedings of the International Conference on Automated
  Planning and Scheduling}, volume~31, pages 510--518, 2021{\natexlab{a}}.

\bibitem[Zhao et~al.(2022)Zhao, Pan, Choromanski, Jain, and
  Sindhwani]{zhao2022implicit}
Yunfan Zhao, Qingkai Pan, Krzysztof Choromanski, Deepali Jain, and Vikas
  Sindhwani.
\newblock Implicit two-tower policies.
\newblock \emph{arXiv preprint arXiv:2208.01191}, 2022.

\bibitem[Sutton(1988)]{sutton1988learning}
Richard~S Sutton.
\newblock Learning to predict by the methods of temporal differences.
\newblock \emph{Machine learning}, 3:\penalty0 9--44, 1988.

\bibitem[Janner et~al.(2021{\natexlab{a}})Janner, Li, and
  Levine]{janner2021offline}
Michael Janner, Qiyang Li, and Sergey Levine.
\newblock Offline reinforcement learning as one big sequence modeling problem.
\newblock \emph{Advances in neural information processing systems},
  34:\penalty0 1273--1286, 2021{\natexlab{a}}.

\bibitem[Zhang and Zhou(2023)]{zhang2023self}
Dan Zhang and Fangfang Zhou.
\newblock Self-supervised image denoising for real-world images with
  context-aware transformer.
\newblock \emph{IEEE Access}, 11:\penalty0 14340--14349, 2023.

\bibitem[Chen et~al.(2020)Chen, Radford, Child, Wu, Jun, Luan, and
  Sutskever]{chen2020generative}
Mark Chen, Alec Radford, Rewon Child, Jeffrey Wu, Heewoo Jun, David Luan, and
  Ilya Sutskever.
\newblock Generative pretraining from pixels.
\newblock In \emph{International conference on machine learning}, pages
  1691--1703. PMLR, 2020.

\bibitem[Brown et~al.(2020)Brown, Mann, Ryder, Subbiah, Kaplan, Dhariwal,
  Neelakantan, Shyam, Sastry, Askell, et~al.]{brown2020language}
Tom Brown, Benjamin Mann, Nick Ryder, Melanie Subbiah, Jared~D Kaplan, Prafulla
  Dhariwal, Arvind Neelakantan, Pranav Shyam, Girish Sastry, Amanda Askell,
  et~al.
\newblock Language models are few-shot learners.
\newblock \emph{Advances in neural information processing systems},
  33:\penalty0 1877--1901, 2020.

\bibitem[Radford et~al.(2018)Radford, Narasimhan, Salimans, Sutskever,
  et~al.]{radford2018improving}
Alec Radford, Karthik Narasimhan, Tim Salimans, Ilya Sutskever, et~al.
\newblock Improving language understanding by generative pre-training.
\newblock 2018.

\bibitem[Janner et~al.(2021{\natexlab{b}})Janner, Li, and
  Levine]{Janner2021ReinforcementLA}
Michael Janner, Qiyang Li, and Sergey Levine.
\newblock Reinforcement learning as one big sequence modeling problem.
\newblock In \emph{Neural Information Processing Systems}, 2021{\natexlab{b}}.

\bibitem[Chen et~al.(2021{\natexlab{b}})Chen, Lu, Rajeswaran, Lee, Grover,
  Laskin, Abbeel, Srinivas, and Mordatch]{chen2021decision}
Lili Chen, Kevin Lu, Aravind Rajeswaran, Kimin Lee, Aditya Grover, Misha
  Laskin, Pieter Abbeel, Aravind Srinivas, and Igor Mordatch.
\newblock Decision transformer: Reinforcement learning via sequence modeling.
\newblock \emph{Advances in neural information processing systems},
  34:\penalty0 15084--15097, 2021{\natexlab{b}}.

\bibitem[Vaswani et~al.(2017)Vaswani, Shazeer, Parmar, Uszkoreit, Jones, Gomez,
  Kaiser, and Polosukhin]{vaswani2017attention}
Ashish Vaswani, Noam Shazeer, Niki Parmar, Jakob Uszkoreit, Llion Jones,
  Aidan~N Gomez, {\L}ukasz Kaiser, and Illia Polosukhin.
\newblock Attention is all you need.
\newblock \emph{Advances in neural information processing systems}, 30, 2017.

\bibitem[Emmons et~al.(2021)Emmons, Eysenbach, Kostrikov, and Levine]{rvs}
Scott Emmons, Benjamin Eysenbach, Ilya Kostrikov, and Sergey Levine.
\newblock Rvs: What is essential for offline rl via supervised learning?
\newblock \emph{arXiv preprint arXiv:2112.10751}, 2021.

\bibitem[Schmidhuber(2019)]{Schmidhuber2019ReinforcementLU}
Juergen Schmidhuber.
\newblock Reinforcement learning upside down: Don't predict rewards - just map
  them to actions.
\newblock \emph{ArXiv}, abs/1912.02875, 2019.

\bibitem[Srivastava et~al.(2019)Srivastava, Shyam, Mutz, Jaśkowski, and
  Schmidhuber]{Srivastava2019TrainingAU}
Rupesh~Kumar Srivastava, Pranav Shyam, Filipe~Wall Mutz, Wojciech Jaśkowski,
  and J{\"u}rgen Schmidhuber.
\newblock Training agents using upside-down reinforcement learning.
\newblock \emph{ArXiv}, abs/1912.02877, 2019.

\bibitem[Jiang and Li(2016)]{Jiang2015DoublyRO}
Nan Jiang and Lihong Li.
\newblock Doubly robust off-policy value evaluation for reinforcement learning.
\newblock In \emph{International Conference on Machine Learning}, pages
  652--661. PMLR, 2016.

\bibitem[Precup et~al.(2000)Precup, Sutton, and Singh]{Precup2000EligibilityTF}
Doina Precup, Richard~S. Sutton, and Satinder Singh.
\newblock Eligibility traces for off-policy policy evaluation.
\newblock In \emph{International Conference on Machine Learning}, 2000.

\bibitem[Hirano et~al.(2003)Hirano, Imbens, and Ridder]{hirano2003efficient}
Keisuke Hirano, Guido~W Imbens, and Geert Ridder.
\newblock Efficient estimation of average treatment effects using the estimated
  propensity score.
\newblock \emph{Econometrica}, 71\penalty0 (4):\penalty0 1161--1189, 2003.

\bibitem[Murphy et~al.(2001)Murphy, van~der Laan, Robins, and
  Group]{murphy2001marginal}
Susan~A Murphy, Mark~J van~der Laan, James~M Robins, and Conduct Problems
  Prevention~Research Group.
\newblock Marginal mean models for dynamic regimes.
\newblock \emph{Journal of the American Statistical Association}, 96\penalty0
  (456):\penalty0 1410--1423, 2001.

\bibitem[Wang et~al.(2017)Wang, Agarwal, and Dud{\i}k]{wang2017optimal}
Yu-Xiang Wang, Alekh Agarwal, and Miroslav Dud{\i}k.
\newblock Optimal and adaptive off-policy evaluation in contextual bandits.
\newblock In \emph{International Conference on Machine Learning}, pages
  3589--3597. PMLR, 2017.

\bibitem[Gottesman et~al.(2019)Gottesman, Liu, Sussex, Brunskill, and
  Doshi-Velez]{gottesman2019combining}
Omer Gottesman, Yao Liu, Scott Sussex, Emma Brunskill, and Finale Doshi-Velez.
\newblock Combining parametric and nonparametric models for off-policy
  evaluation.
\newblock In \emph{International Conference on Machine Learning}, pages
  2366--2375. PMLR, 2019.

\bibitem[Hanna et~al.(2019)Hanna, Niekum, and Stone]{hanna2019importance}
Josiah Hanna, Scott Niekum, and Peter Stone.
\newblock Importance sampling policy evaluation with an estimated behavior
  policy.
\newblock In \emph{International Conference on Machine Learning}, pages
  2605--2613. PMLR, 2019.

\bibitem[Hanna and Stone(2018)]{hanna2018towards}
Josiah~P Hanna and Peter Stone.
\newblock Towards a data efficient off-policy policy gradient.
\newblock In \emph{AAAI Spring Symposia}, 2018.

\bibitem[Chen et~al.(2023{\natexlab{a}})Chen, Zhang, Riem, Adam, Bastian, and
  Lan]{chen2023explainable}
Jingdi Chen, Lei Zhang, Joseph Riem, Gina Adam, Nathaniel~D Bastian, and Tian
  Lan.
\newblock Explainable learning-based intrusion detection supported by
  memristors.
\newblock In \emph{2023 IEEE Conference on Artificial Intelligence (CAI)},
  pages 195--196. IEEE, 2023{\natexlab{a}}.

\bibitem[Mei et~al.(2023{\natexlab{a}})Mei, Zhou, and Lan]{mei2023remix}
Yongsheng Mei, Hanhan Zhou, and Tian Lan.
\newblock Remix: Regret minimization for monotonic value function factorization
  in multiagent reinforcement learning.
\newblock \emph{arXiv preprint arXiv:2302.05593}, 2023{\natexlab{a}}.

\bibitem[Li et~al.(2015)Li, Munos, and Szepesv{\'a}ri]{li2015toward}
Lihong Li, R{\'e}mi Munos, and Csaba Szepesv{\'a}ri.
\newblock Toward minimax off-policy value estimation.
\newblock In \emph{Artificial Intelligence and Statistics}, pages 608--616.
  PMLR, 2015.

\bibitem[He et~al.(2023)He, Han, Su, Han, Zou, and Miao]{he2023robust}
Sihong He, Songyang Han, Sanbao Su, Shuo Han, Shaofeng Zou, and Fei Miao.
\newblock Robust multi-agent reinforcement learning with state uncertainty.
\newblock \emph{Transactions on Machine Learning Research}, 2023.

\bibitem[He et~al.(2020)He, Pepin, Wang, Zhang, and Miao]{he2020data}
Sihong He, Lynn Pepin, Guang Wang, Desheng Zhang, and Fei Miao.
\newblock Data-driven distributionally robust electric vehicle balancing for
  mobility-on-demand systems under demand and supply uncertainties.
\newblock In \emph{2020 IEEE/RSJ International Conference on Intelligent Robots
  and Systems (IROS)}, pages 2165--2172. IEEE, 2020.

\bibitem[Narita et~al.(2019)Narita, Yasui, and Yata]{narita2019efficient}
Yusuke Narita, Shota Yasui, and Kohei Yata.
\newblock Efficient counterfactual learning from bandit feedback.
\newblock In \emph{Proceedings of the AAAI Conference on Artificial
  Intelligence}, volume~33, pages 4634--4641, 2019.

\bibitem[Delyon and Portier(2016)]{delyon2016integral}
Bernard Delyon and Fran{\c{c}}ois Portier.
\newblock Integral approximation by kernel smoothing.
\newblock 2016.

\bibitem[Kumar et~al.(2020)Kumar, Zhou, Tucker, and Levine]{cql}
Aviral Kumar, Aurick Zhou, George Tucker, and Sergey Levine.
\newblock Conservative q-learning for offline reinforcement learning.
\newblock \emph{Advances in Neural Information Processing Systems},
  33:\penalty0 1179--1191, 2020.

\bibitem[Kumar et~al.(2019)Kumar, Fu, Soh, Tucker, and Levine]{bear}
Aviral Kumar, Justin Fu, Matthew Soh, George Tucker, and Sergey Levine.
\newblock Stabilizing off-policy q-learning via bootstrapping error reduction.
\newblock \emph{Advances in Neural Information Processing Systems}, 32, 2019.

\bibitem[Wu et~al.(2021)Wu, Zhai, Srivastava, Susskind, Zhang, Salakhutdinov,
  and Goh]{uwac}
Yue Wu, Shuangfei Zhai, Nitish Srivastava, Joshua Susskind, Jian Zhang, Ruslan
  Salakhutdinov, and Hanlin Goh.
\newblock Uncertainty weighted actor-critic for offline reinforcement learning.
\newblock \emph{arXiv preprint arXiv:2105.08140}, 2021.

\bibitem[Wu et~al.(2019)Wu, Tucker, and Nachum]{brac}
Yifan Wu, George Tucker, and Ofir Nachum.
\newblock Behavior regularized offline reinforcement learning.
\newblock \emph{arXiv preprint arXiv:1911.11361}, 2019.

\bibitem[Kostrikov et~al.(2021)Kostrikov, Nair, and Levine]{iql}
Ilya Kostrikov, Ashvin Nair, and Sergey Levine.
\newblock Offline reinforcement learning with implicit q-learning.
\newblock \emph{arXiv preprint arXiv:2110.06169}, 2021.

\bibitem[Puterman(2014)]{puterman2014markov}
Martin~L Puterman.
\newblock \emph{Markov decision processes: discrete stochastic dynamic
  programming}.
\newblock John Wiley \& Sons, 2014.

\bibitem[Mei et~al.(2022)Mei, Lan, Imani, and Subramaniam]{mei2022bayesian}
Yongsheng Mei, Tian Lan, Mahdi Imani, and Suresh Subramaniam.
\newblock A bayesian optimization framework for finding local optima in
  expensive multi-modal functions.
\newblock \emph{arXiv preprint arXiv:2210.06635}, 2022.

\bibitem[Agarwal et~al.(2022)Agarwal, Aggarwal, and Lan]{agarwal2022multi}
Mridul Agarwal, Vaneet Aggarwal, and Tian Lan.
\newblock Multi-objective reinforcement learning with non-linear scalarization.
\newblock In \emph{Proceedings of the 21st International Conference on
  Autonomous Agents and Multiagent Systems}, pages 9--17, 2022.

\bibitem[Ma(2022)]{ma2022traffic}
Xiaobo Ma.
\newblock \emph{Traffic Performance Evaluation Using Statistical and Machine
  Learning Methods}.
\newblock PhD thesis, The University of Arizona, 2022.

\bibitem[Henmi et~al.(2007)Henmi, Yoshida, and Eguchi]{henmi2007importance}
Masayuki Henmi, Ryo Yoshida, and Shinto Eguchi.
\newblock Importance sampling via the estimated sampler.
\newblock \emph{Biometrika}, 94\penalty0 (4):\penalty0 985--991, 2007.

\bibitem[Kallus and Uehara(2020)]{kallus2020statistically}
Nathan Kallus and Masatoshi Uehara.
\newblock Statistically efficient off-policy policy gradients.
\newblock In \emph{International Conference on Machine Learning}, pages
  5089--5100. PMLR, 2020.

\bibitem[Williams(1992)]{williams1992simple}
Ronald~J Williams.
\newblock Simple statistical gradient-following algorithms for connectionist
  reinforcement learning.
\newblock \emph{Reinforcement learning}, pages 5--32, 1992.

\bibitem[Chen and Jiang(2019)]{chen2019information}
Jinglin Chen and Nan Jiang.
\newblock Information-theoretic considerations in batch reinforcement learning.
\newblock In \emph{International Conference on Machine Learning}, pages
  1042--1051. PMLR, 2019.

\bibitem[Zhou et~al.(2022)Zhou, Lan, and Aggarwal]{zhou2022pac}
Hanhan Zhou, Tian Lan, and Vaneet Aggarwal.
\newblock Pac: Assisted value factorization with counterfactual predictions in
  multi-agent reinforcement learning.
\newblock In \emph{Advances in Neural Information Processing Systems},
  volume~35, pages 15757--15769. Curran Associates, Inc., 2022.

\bibitem[Chen et~al.(2022)Chen, Chen, Lan, and Aggarwal]{chen2022scalable}
Jiayu Chen, Jingdi Chen, Tian Lan, and Vaneet Aggarwal.
\newblock Scalable multi-agent covering option discovery based on kronecker
  graphs.
\newblock \emph{Advances in Neural Information Processing Systems},
  35:\penalty0 30406--30418, 2022.

\bibitem[Mei et~al.(2023{\natexlab{b}})Mei, Zhou, Lan, Venkataramani, and
  Wei]{mei2023mac}
Yongsheng Mei, Hanhan Zhou, Tian Lan, Guru Venkataramani, and Peng Wei.
\newblock Mac-po: Multi-agent experience replay via collective priority
  optimization.
\newblock In \emph{Proceedings of the 2023 International Conference on
  Autonomous Agents and Multiagent Systems}, pages 466--475,
  2023{\natexlab{b}}.

\bibitem[Zhou et~al.(2023)Zhou, Lan, and Aggarwal]{zhou2022value}
Hanhan Zhou, Tian Lan, and Vaneet Aggarwal.
\newblock Value functions factorization with latent state information sharing
  in decentralized multi-agent policy gradients.
\newblock \emph{IEEE Transactions on Emerging Topics in Computational
  Intelligence}, pages 1--11, 2023.
\newblock \doi{10.1109/TETCI.2023.3293193}.

\bibitem[Chen et~al.(2021{\natexlab{c}})Chen, Wang, and Lan]{chen2021bringing}
Jingdi Chen, Yimeng Wang, and Tian Lan.
\newblock Bringing fairness to actor-critic reinforcement learning for network
  utility optimization.
\newblock In \emph{IEEE INFOCOM 2021-IEEE Conference on Computer
  Communications}, pages 1--10. IEEE, 2021{\natexlab{c}}.

\bibitem[Chen et~al.(2023{\natexlab{b}})Chen, Lan, and
  Aggarwal]{chen2023option}
Jiayu Chen, Tian Lan, and Vaneet Aggarwal.
\newblock Option-aware adversarial inverse reinforcement learning for robotic
  control.
\newblock In \emph{2023 IEEE International Conference on Robotics and
  Automation (ICRA)}, pages 5902--5908. IEEE, 2023{\natexlab{b}}.

\bibitem[Silver et~al.(2014)Silver, Lever, Heess, Degris, Wierstra, and
  Riedmiller]{silver2014deterministic}
David Silver, Guy Lever, Nicolas Heess, Thomas Degris, Daan Wierstra, and
  Martin Riedmiller.
\newblock Deterministic policy gradient algorithms.
\newblock In \emph{International conference on machine learning}, pages
  387--395. Pmlr, 2014.

\bibitem[Levine and Koltun(2013)]{levine2013guided}
Sergey Levine and Vladlen Koltun.
\newblock Guided policy search.
\newblock In \emph{International conference on machine learning}, pages 1--9.
  PMLR, 2013.

\bibitem[Elmachtoub et~al.(2023)Elmachtoub, Gupta, and
  Zhao]{elmachtoub2023balanced}
Adam Elmachtoub, Vishal Gupta, and Yunfan Zhao.
\newblock Balanced off-policy evaluation for personalized pricing.
\newblock In \emph{International Conference on Artificial Intelligence and
  Statistics}, pages 10901--10917. PMLR, 2023.

\bibitem[Jie and Abbeel(2010)]{jie2010connection}
Tang Jie and Pieter Abbeel.
\newblock On a connection between importance sampling and the likelihood ratio
  policy gradient.
\newblock \emph{Advances in Neural Information Processing Systems}, 23, 2010.

\bibitem[Thomas and Brunskill(2016)]{Thomas2016DataEfficientOP}
Philip~S. Thomas and Emma Brunskill.
\newblock Data-efficient off-policy policy evaluation for reinforcement
  learning.
\newblock \emph{ArXiv}, abs/1604.00923, 2016.

\bibitem[Deisenroth et~al.(2013)Deisenroth, Neumann, Peters,
  et~al.]{deisenroth2013survey}
Marc~Peter Deisenroth, Gerhard Neumann, Jan Peters, et~al.
\newblock A survey on policy search for robotics.
\newblock \emph{Foundations and Trends{\textregistered} in Robotics},
  2\penalty0 (1--2):\penalty0 1--142, 2013.

\bibitem[Rasmussen and Ghahramani(2003)]{rasmussen2003bayesian}
Carl~Edward Rasmussen and Zoubin Ghahramani.
\newblock Bayesian monte carlo.
\newblock \emph{Advances in neural information processing systems}, pages
  505--512, 2003.

\bibitem[Fu et~al.(2020)Fu, Kumar, Nachum, Tucker, and Levine]{fu2020d4rl}
Justin Fu, Aviral Kumar, Ofir Nachum, George Tucker, and Sergey Levine.
\newblock D4rl: Datasets for deep data-driven reinforcement learning.
\newblock \emph{arXiv preprint arXiv:2004.07219}, 2020.

\bibitem[Paster et~al.(2022)Paster, McIlraith, and Ba]{paster2022you}
Keiran Paster, Sheila McIlraith, and Jimmy Ba.
\newblock You can't count on luck: Why decision transformers fail in stochastic
  environments.
\newblock \emph{arXiv preprint arXiv:2205.15967}, 2022.

\end{thebibliography}
